\documentclass[lettersize,journal]{IEEEtran}
\usepackage{amsmath,amsfonts}
\usepackage{algorithmic}
\usepackage{algorithm}
\usepackage{array}
\usepackage[caption=false,font=normalsize,labelfont=sf,textfont=sf]{subfig}
\usepackage{textcomp}
\usepackage{stfloats}
\usepackage{url}
\usepackage{verbatim}
\usepackage{graphicx}
\usepackage{cite}

\usepackage{multirow}
\usepackage{subfig}
\usepackage{color}
\usepackage{soul} 
\usepackage{xcolor}
\usepackage{makecell}
\usepackage{bm}
\usepackage{mathrsfs}
\usepackage{balance}

\usepackage{hyperref} 
\hypersetup{
hidelinks,
linkcolor=black,
citecolor=black,
urlcolor = black}

\begin{document}

\title{Multi-Evidence based Fact Verification via A Confidential Graph Neural Network}

\author{Yuqing Lan, Zhenghao Liu, Yu Gu\thanks{(Corresponding author: Zhenghao Liu, Yu Gu.)}, Xiaoyuan Yi, Xiaohua Li, Liner Yang, Ge Yu, ~\IEEEmembership{Member,~IEEE, }
        
\thanks{Yuqing Lan, Zhenghao Liu, Yu Gu, Xiaohua Li and Ge Yu are with the School of Computer Science and
Engineering, Northeastern University, Shenyang, Liaoning 110819, China. Email: lanyuqing@stumail.neu.edu.cn, {liuzhenghao, Xiaohua Li, guyu, yuge}@mail.neu.edu.cn. Xiaoyuan Yi is with Microsoft Research Asia, Beijing, China. E-mail: xiaoyuanyi@microsoft.com. Liner Yang is with the National Language Resources Monitoring and Research Center for Print Media, Beijing Language and Culture University, Beijing,
China. Email: lineryang@gmail.com.
}
}



\maketitle

\begin{abstract}
Fact verification tasks aim to identify the integrity of textual contents according to the truthful corpus. Existing fact verification models usually build a fully connected reasoning graph, which regards claim-evidence pairs as nodes and connects them with edges. They employ the graph to propagate the semantics of the nodes. Nevertheless, the noisy nodes usually propagate their semantics via the edges of the reasoning graph, which misleads the semantic representations of other nodes and amplifies the noise signals. To mitigate the propagation of noisy semantic information, we introduce a Confidential Graph Attention Network (CO-GAT), which proposes a node masking mechanism for modeling the nodes. Specifically, CO-GAT calculates the node confidence score by estimating the relevance between the claim and evidence pieces. Then, the node masking mechanism uses the node confidence scores to control the noise information flow from the vanilla node to the other graph nodes. CO-GAT achieves a 73.59\% FEVER score on the FEVER dataset and shows the generalization ability by broadening the effectiveness to the science-specific domain.
\end{abstract}

\section{Introduction}
\IEEEPARstart
{R}{ecently}, the proliferation of fake news and rumors on the internet has become increasingly prevalent~\cite{cheng2021causal,zafarani2019fake,DBLP:journals/tois/YangSZLC17}, which potentially misguides the public, induces panic, and harms the reputation of individuals, organizations, or companies. Automatic fact-checking~\cite{vosoughi2018spread,hassan2015detecting,vlachos2014fact} plays a pivotal role in enhancing information verification efficiency, reducing the workload on human labor, and mitigating the risks associated with human biases. Therefore, the fact verification systems~\cite{guo2022survey,wadden-etal-2020-fact, DBLP:conf/emnlp/JiangBZD0B20, DBLP:conf/acl/ParkMKZH22,ma2019sentence,wan2021dqn} are proposed to automatically verify the truthfulness and credibility of the given claim with the trust-worthy corpus~\cite{DBLP:journals/csur/BekoulisPD23,augenstein-etal-2019-multifc,schuster2021get}, such as Wikipedia~\cite{thorne2018fact,schuster2019towards} and knowledge graph~\cite{kim-etal-2023-factkg}.

\begin{figure}[t]
    \centering 

    \includegraphics[width=3.0in]{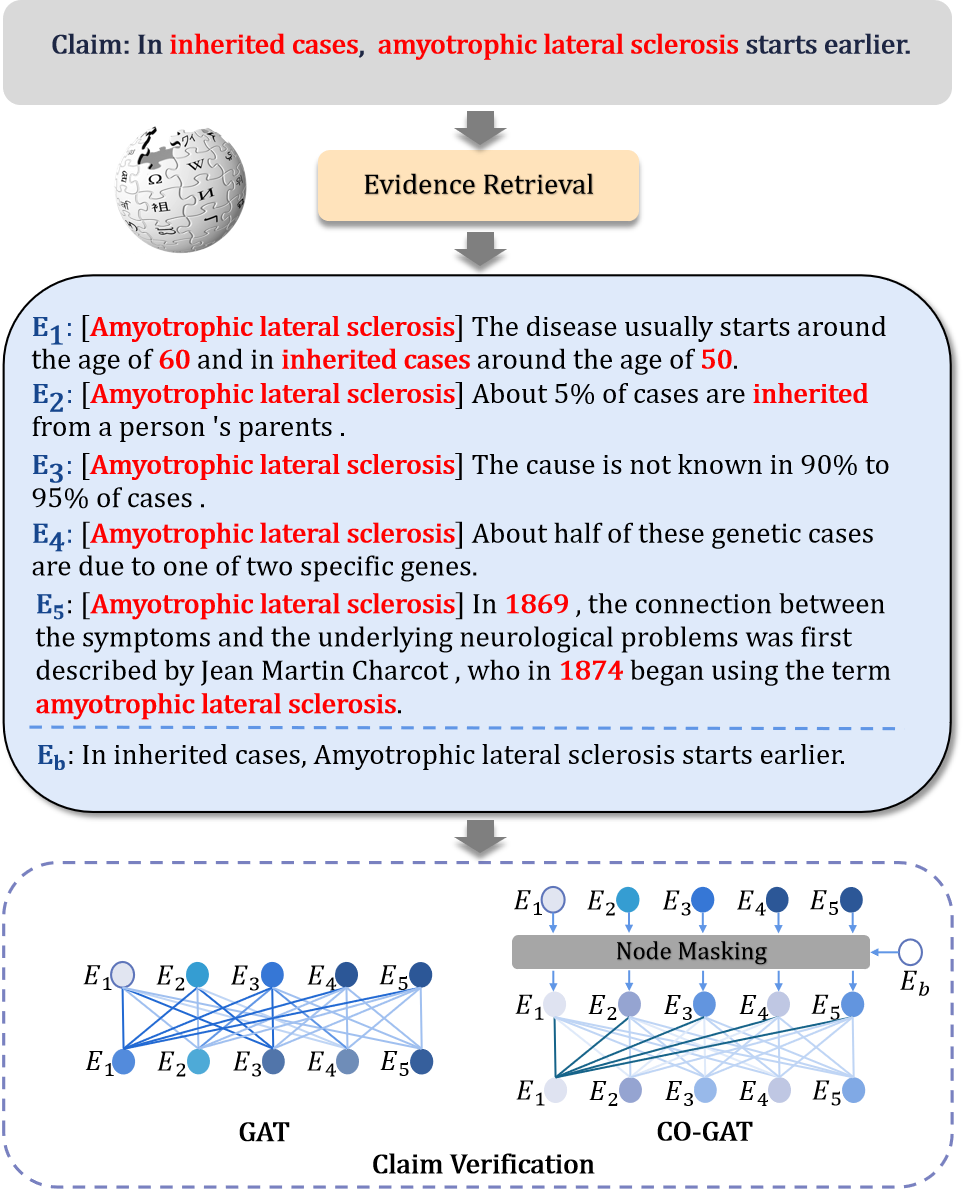}
    \caption{An Example of Fact Verification System. $E_1$ represents the golden evidence of the given claim. The remained evidence nodes indicates that the claim and evidence piece are irrelevant. $E_b$ represents the blank node. The given claim and each piece of evidence are concatenated as graph nodes. Nodes are connected via edges, forming a fully connected graph for claim verification reasoning.
    }
    \label{fig:intro}
\end{figure}

Existing work usually employs a three-step pipeline model~\cite{fajcik-etal-2023-claim,yin2018twowingos,chen2017reading} to verify the given claims. Specifically, they design a document retrieval model~\cite{DBLP:conf/emnlp/Aly022,nie2019combining} to search relevant documents for the given claim. Subsequently, they utilize a sentence retrieval model~\cite{hanselowski2018ukp,devlin2019bert} to select multiple relevant evidence sentences from the retrieved documents to validate the given claim. Lots of previous work~\cite{DBLP:conf/acl/LiuXSL20,zhou2019gear,chen2022loren,DBLP:conf/coling/ParkLJKKN22} pays more attention to the claim verification step, which conducts multi-evidence reasoning model to verify the integrity of the claim. However, the retrieval models inevitably involve noise information and stimulate the claim verification models to exhaust their efforts to filter out the unrelated evidence pieces. \IEEEpubidadjcol
Most claim verification models~\cite{zhou2019gear,DBLP:conf/acl/LiuXSL20,yoneda2018ucl,si2021topic} employ the Graph Attention Network (GAT)~\cite{velivckovic2018graph} to conduct multi-evidence reasoning. They regard each claim-evidence pair as a node and establish full connections among the graph nodes to facilitate the propagation of semantics among different evidence pieces. Such an attention mechanism establishes deep interactions among nodes, but provides opportunities for semantic information propagation of noisy nodes, leading to an additional noise channel during the node representation learning.

This paper introduces the Confidential Graph Neural Network Reasoning Model (CO-GAT)\footnote{The code is available at \url{https://github.com/NEUIR/CO-GAT}}. As shown in Fig.~\ref{fig:intro}, we design an additional node representation masking mechanism before the graph reasoning modeling, which controls the evidence information flow into the graph reasoning model. Specifically, 
CO-GAT learns the node confidence by estimating the claim-evidence relevance. It integrates the semantic information of blank nodes based on confidence scores to erase noise information of the vanilla node, thereby controlling the flow of evidence information and preventing the propagation of noise signals.
Finally, these masked node representations are fed into the graph attention neural network for claim verification, alleviating the unnecessary noise propagation from the evidence pieces unrelated to the given claim.

Our experiments demonstrate CO-GAT's effectiveness in identifying factual agreements on the FEVER~\cite{thorne2018fact} and SCIFACT~\cite{wadden-etal-2020-fact} datasets. CO-GAT prefers to classify claims as ``NOT ENOUGH INFO'' (NEI) when the model uncertainty is high or the model prediction is incorrect, which thrives on insufficient evidence for fact verification prediction.
Our studies indicate the node masking mechanism effectively erases the noise information of nodes, preventing noise information propagation in the reasoning graph. This leads to a more concentrated edge attention distribution while a more scattered node attention distribution.

In Section~\ref{related work}, we discuss the related work of the fact verification tasks. Section~\ref{method} introduces the methodology of CO-GAT. The experimental details are described in Section~\ref{experimental}, followed by the presentation of experimental results in Section~\ref{result}. Lastly, Section~\ref{conclusion} concludes this paper.

\section{Related Work}\label{related work}
The fact verification task aims to develop automatic fact verification systems, which check the veracity of the claims by retrieving evidence pieces from the trustworthy corpus. It has numerous applications, including question answering~\cite{DBLP:conf/emnlp/WangZLZYZ22} and abstract summarization~\cite{zhang-etal-2020-optimizing}. Current fact verification models commonly follow a three-step pipeline system~\cite{chen2017reading}, which includes document retrieval, sentence retrieval, and claim verification. Most fact verification systems follow the document retrieval model and the sentence retrieval model of previous work~\cite{nie2019combining, hanselowski2018ukp, DBLP:conf/acl/LiuXSL20} and primarily concentrate on the claim verification step. These claim verification models can be grouped into two categories: the sequential models and the graph reasoning based models.

\subsection{Sequential Models}
Previous research on the fact verification task adapts existing Natural Language Inference (NLI) models to predict the label of the given claim~\cite{nie2019combining,parikh2016decomposable,radford2018improving,chen2017enhanced,ghaeini2018dr,peters2018deep,li2019several,hanselowski2018ukp,hidey-diab-2018-team}. The NLI task aims to classify the relationship between a pair of premises and hypotheses as entailment, contradiction, or neutral, similar to the fact verification task. One of the most commonly used models is the Enhanced Sequential Inference Model (ESIM)~\cite{chen-etal-2017-enhanced} and its variants.

Recently, pre-trained language models (PLMs)~\cite{liu2019roberta,devlin2019bert,lewis2020bart,raffel2020exploring, NEURIPS2019_dc6a7e65} have shown their strong effectiveness in fact verification~\cite{DBLP:conf/ecir/SoleimaniMW20,jiang2021exploring}. Some researchers directly use language models to verify claims without any evidence~\cite{lee2021towards}. Other researchers use the Retrieval-augmented generation (RAG) methods for fact verification tasks~\cite{DBLP:conf/iclr/0002IWXJ000023,DBLP:conf/nips/LewisPPPKGKLYR020}. They use a retriever to retrieve claim-related evidence, and then integrate the claim and evidence knowledge into the generation task to verify the claim. Then, some researchers use data augmentation methods to train on the generated datasets based on the FEVER dataset to improve the effectiveness of claim verification~\cite{DBLP:conf/cikm/LeeWKLPJ21,pan-etal-2021-Zero-shot-FV}. Most fact verification systems use pre-trained language models to encode the claim and the evidence~\cite{10.1007/978-3-030-45442-5_45, DBLP:conf/acl/KruengkraiYW21}. Then they combine attention mechanisms~\cite{DBLP:conf/acl/KruengkraiYW21,pradeep2021scientific} or some neural network architectures~\cite{10.1007/978-3-030-45442-5_45,xu2023counterfactual} to construct classification layers for label prediction.

In addition, some claim verification models are formulated based on logical rules of language.
LOREN~\cite{chen2022loren} decomposes the claim into a series of phrases and then uses the logical aggregation rules to validate the phrases and predict the claim label. ProoFVer~\cite{DBLP:journals/tacl/Krishna0022} uses a seq2seq model to generate a natural logic proof to verify the claim. CLEVER~\cite{xu2023counterfactual} utilizes counterfactual theory to achieve claim verification by mitigating bias introduced from solely relying on claims for prediction during the inference stage.

\subsection{Graph Reasoning based Models}
Other works adopt graph-based modeling methods for multi-evidence reasoning.
An early attempt to use a graph-based reasoning model for fact verification is GEAR~\cite{zhou2019gear}, which regards the evidence-claim pair as a graph node to construct a fully connected evidence graph. It proposes an evidence reasoning network, similar to a graph attention neural network~\cite{velivckovic2018graph} to propagate evidence information.
Then, KGAT~\cite{DBLP:conf/acl/LiuXSL20} proposes a fine-grained fact verification model based on a kernel graph attention neural network. It utilizes both token-level and sentence-level kernel attention mechanisms to capture fine-grained evidence information for claim verification. EvidenceNet~\cite{Chen-evidencenet} utilizes the evidence fusion network to capture different levels of global contextual information filtered by the gating mechanism. ICMI~\cite{wang-etal-2022-imci} is inspired by the multi-relational graph convolutional network and then uses dual evidence fusion graphs to capture multi-view information within and between documents, thereby solving the problem of multi-hop fact verification. CGAT~\cite{barik2022incorporating} constructs a phrase-level graph by introducing a knowledge graph to obtain a node representation with detailed semantic information and then uses the graph attention network (GAT) model for inference. 
FACTKG~\cite{kim-etal-2023-factkg} creates a dataset by using a knowledge graph, which provides an evidence graph composed of entities and entity relations. It uses GEAR~\cite{zhou2019gear} for factual verification reasoning. 

In addition, some work introduces semantic graphs for factual verification. It uses the semantic role labeling (SRL) toolkit to parse the evidence sentences into $n$ structural relation triples as the nodes.
DREAM~\cite{zhong-etal-2020-reasoning} constructs semantic graphs for claim and evidence, which encodes them using the XLNET model~\cite{NEURIPS2019_dc6a7e65}. Meanwhile, it uses graph convolutional network (GNN)~\cite{kipf2016semi} and the GAT model to propagate and aggregate evidence information. GLAF~\cite{DBLP:conf/coling/MaL0C22} constructs an evidence graph based on triple-level nodes. It combines the local fission reasoning layer and global evidence aggregation layer to establish logical relations between clues, share information, and exchange evidence clues for claim verification. 
RoEG~\cite{chen2021entity} uses an entity recognizer to extract entities from the evidence. It then constructs an entity graph to capture fine-grained evidence features and semantic relations. SISER~\cite{DBLP:conf/coling/ParkLJKKN22} incorporates sentence-level graph reasoning, sequence reasoning, and semantic graph reasoning to verify the claim.

\section{Methodology}\label{method}
\begin{figure}[t]
    \centering 

    \includegraphics[width=3.5in]{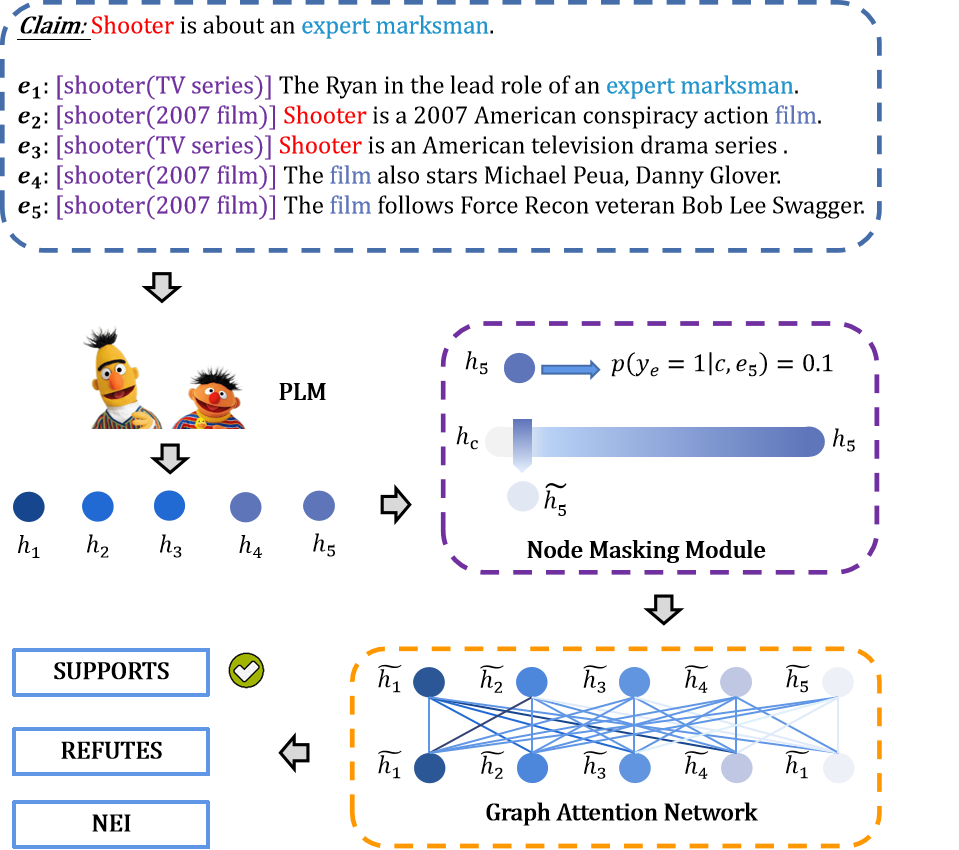}
    \caption{Illustration of CO-GAT. It is mainly divided into three steps. Firstly, the pre-trained language models are used to encode the graph nodes, which are claim-evidence pairs. Secondly, the node masking mechanism is utilized to erase noise information based on the confidence score (CO-SCO) before graph reasoning. Finally, we use a fully connected graph to propagate the denoised evidence information among nodes and aggregate node information for predicting the claim label. 
    }
    \label{fig:model}
\end{figure}
In this section, we provide an overview of the Confidential Graph Attention Network (CO-GAT) framework and its application in the fact verification task.

CO-GAT aims to predict the factual label $y_c$ of the given claim $c$ with the graph neural network (Sec.~\ref{model: gat}). Similar to the previous work~\cite{zhou2019gear,DBLP:conf/acl/LiuXSL20,DBLP:conf/coling/ParkLJKKN22}, we regard the claim-evidence pair $<c, e_p>$ as the node $n_p$. We construct a fully connected graph $G$ using $l$ nodes, $N=\{n_1, \dots, n_p, \dots, n_l\}$. As shown in Fig.~\ref{fig:model}, the CO-GAT model initially employs a pre-trained language model to encode the graph nodes. Subsequently, CO-GAT utilizes the node masking mechanism (Sec.~\ref{model: Evidence ablation module}) to erase the noise semantic information before feeding the node representations into the graph attention neural network. Finally, we use the multi-task modeling methods (Sec.~\ref{model: training}) to train CO-GAT.

\subsection{Graph Reasoning for Fact Verification}\label{model: gat}
The CO-GAT model uses the retrieved evidence $E=\{e_1,e_2,\dots,e_i\}$ and the given claim $c$ to construct the evidence reasoning graph $G$ for the fact verification task.

\textbf{Node Encoding.} The node representation is initialized by feeding the concatenated sequence to the pre-trained language models, such as ELECTRA~\cite{clark2020electra} and RoBERTa~\cite{liu2019roberta}.
\begin{equation}\label{representaion}
    h_p=\text{PLM}(\text{[CLS]}\circ c \circ e_p \circ \text{[SEP]}),
\end{equation}
where $\circ$ denotes the concatenation operation. As shown in Fig.~\ref{fig:model}, $e_p$ contains the retrieved evidence sentence and the document (Wiki) title. Both ``[CLS]'' and ``[SEP]'' are special tokens in the pre-trained language models, such as ELECTRA and RoBERTa. We take the last hidden state of the ``[CLS]'' token as the initial node representation $h_p$ of the node $n_p$.

\textbf{Graph Reasoning.} Following the previous work~\cite{zhou2019gear,velivckovic2018graph}, the evidence reasoning graph uses the edge attention mechanism to propagate evidence information and the node attention mechanism to aggregate node information for claim prediction.

For the edge attention mechanism, each node receives attention from neighbor nodes and uses this attention score to aggregate semantic information from neighbor nodes to update the node representation. CO-GAT utilizes the multi-head attention mechanism~\cite{vaswani2017attention} to construct the edge attention mechanism. Specifically, the scaled dot product attention model is used to calculate the edge attention: 
\begin{equation}
  \alpha_i^{q \rightarrow p} 	= \text{Softmax}_q(\frac{h_p\cdot W_i^Q \times h_q\cdot W_i^K}{\sqrt{d_k}}),\label{equation:edge-Attention}
\end{equation}
\begin{equation}
  d_k=d_m/\#head\label{equation:dk},
\end{equation}
where $\alpha_i^{q \rightarrow p}$ represents the $i$-th head edge attention that the node $n_q$ receives from the node $n_p$. The weight matrix $W_i^Q, W_i^K \in \mathcal{R}^{d_m \times d_k}$. $d_k$ represents the dimension of each head, which is calculated by Eq.~\ref{equation:dk}. $d_m$ denotes the hidden size of the pre-trained language models. $\#head$ represents the number of attention heads.

Subsequently, the node $n_p$ uses the edge attention score $\alpha_i^{q \rightarrow p}$ to aggregate the semantic information from neighbor nodes. CO-GAT concatenates the node representation $v^i (h_p)$ obtained from each attention head as the updated node representation $v (h_p)$:
\begin{equation}
  v^i (h_p)= \sum_{i=1}^l \alpha_i^{q \rightarrow p} \times (h_p \cdot W_i^V)\label{equation:Attention},
\end{equation}
\begin{equation}
    v (h_p)=v^1 (h_p),\dots,v^h (h_p).
\end{equation}
where $W_i^V \in \mathcal{R}^{d_m \times d_k}$.
Through the edge attention mechanism, each node obtains semantic information from the neighbor nodes. Consequently, the node attention mechanism assigns each node with an attention weight to aggregate the semantic information of graph nodes for claim verification.
We use the Eq.~\ref{equation:node-Attention} to calculate the node attention score $\beta_{p}$ of the node $n_p$: 
\begin{equation}
    \beta_{p} = \text{Softmax}_p(\text{Linear} (v (h_p))).\label{equation:node-Attention}
\end{equation}

Then, we employ the node aggregator to gather the semantic information from all graph nodes to get the final hidden state $V$:
\begin{equation}
  V = \sum_{p=1}^l \beta_p \times v (h_p).\label{equation:final_representation}
\end{equation}

\textbf{Label Prediction.} The factual classification label $y_c$ of the given claim $c$ is predicted according to the probability $P(y_c|c,E)$: 
\begin{equation}
   P(y_c|c,E)=\text{Softmax}_{y_c}(\text{Linear} (V)),\label{equation:label}
\end{equation}
where the labels $y_c$ are categorized into three classes: 0, 1, and 2, representing SUPPORTS, REFUTES, and NOT ENOUGH INFO, respectively. These labels signify whether the provided evidence supports, refutes, or lacks sufficient information to predict the claim.

Next, in Sec.~\ref{model: Evidence ablation module}, we introduce a node masking mechanism to erase the noise semantic information before feeding the node representations to graph reasoning. Subsequently, we employ the multi-task modeling method outlined in Section~\ref{model: training} to train the CO-GAT model.

\subsection{Confidential Graph Reasoning of Node Masking}\label{model: Evidence ablation module}
The task of fact verification requires the utilization of multiple pieces of evidence for reasoning. However, the retrieval evidence pieces inevitably contain noise information. 
As the attention layer deepens, the node representation tends to become homogeneous~\cite{clark-etal-2019-bert,brody2021attentive}. Therefore, the noise semantic information is easily captured by other nodes during graph reasoning, consequently affecting the representation learning of the graph nodes.
Therefore, the CO-GAT model proposes a node masking mechanism to erase the noise information from the initial node representations and alleviate the noisy semantic information propagated in the graph neural network. 

The model assesses the confidence score (CO-SCO) of nodes and adjusts the information flow of blank nodes according to this score. This process aims to conduct the denoised node presentation $\widetilde{h}_p$ of the node $n_p$ by adding the representation of blank node to the initial node representations using CO-SCO:
\begin{equation}
  \widetilde{h}_p	= \text{CO-SCO} \times h_p + (1-\text{CO-SCO})\times h_b,\label{equation:cover}
\end{equation}
where $h_b$ is the node representation of the blank node $n_b$. The blank node $n_b$ only contains the claim $c$ without any evidence. Similarity to Eq.~\ref{representaion}, we utilize the same pre-trained language model for encoding the node $n_b$.

The confidence score (CO-SCO) can be calculated:
\begin{equation}\label{equation: confidence}
\begin{split}
    \text{CO-SCO} &=P(y_{e_p}=1|c,e_p) \\
    &=\text{Softmax}_{y_{e_p}=1}(\text{Linear} (h_p)),
\end{split}
\end{equation}
where $\text{CO-SCO}\in[0,1]$. The $y_{e_p}=1$ indicates that the model recognizes the given evidence $e_p$ as the golden evidence for the claim $c$. CO-SCO represents the relevance between the claim $c$ and evidence $e_p$. A higher confidence score indicates that the evidence piece is more effective in supporting or refuting the claim, indicating the necessity to retain more semantic information from the initial node representation.

\subsection{Multi-Task Modeling Methods}\label{model: training}
The CO-GAT model utilizes a multi-task training strategy, in which the final loss function $L$ consists of two loss functions: the cross entropy loss function $L_{fact}$ for claim verification, and the cross entropy loss function $L_{evi}$ for node prediction:
\begin{equation}\label{eq:loss}
L = L_{fact} + L_{evi}.
\end{equation}

\textbf{Claim Verification Loss.} 
 The claim verification model can be trained by minimizing the cross entropy loss with the claim prediction label $y_c$:
\begin{equation}\label{eq:factloss}
L_{fact} = \text{CrossEntropy}(y_{c}^*, P(y_c|c,E)),
\end{equation}
where $y_{c}^*$ is the ground truth verification label for the given claim $c$. The labels $y_{c}^*$ and $y_c$ can categorized into three classes, representing SUPPORTS, REFUTES, and NOT ENOUGH INFO, respectively.

To obtain the prediction label $y_c$, CO-GAT employs the graph attention network as the reasoning model to integrate the evidence information:
\begin{equation}
   P(y_c|c,E)=\text{Softmax}_{y_c}(\text{Linear} ( \widetilde{V})).
\end{equation}

The final hidden state $\widetilde{V}$ of the evidence graph is obtained by Eq.~\ref{equation:final_representation} using the denoised node representation $\widetilde{h}_p$ (Eq.~\ref{equation:cover}):
\begin{equation}
  \widetilde{V} = \sum_{p=1}^l \beta_p \times v (\widetilde{h}_p).
\end{equation}

\textbf{Node Prediction Loss.}
For each node in the evidence graph, we predict its confidential label 
$y_e$. we minimize the node prediction loss $L_{evi}$ to help language models judge the relevance between claims-evidence pair and predict the node confidence score:
\begin{equation}\label{eq:eviloss}
L_{evi} = \text{CrossEntropy}(y_{e_p}^*, P(y_{e_p}|c,E)),
\end{equation}
where $y_{e_p}^*$ denotes the ground truth confidence label for the graph node $n_p$. $y_{e_p}^*$ and $y_{e_p}$ can be categorized into two groups: 0 and 1, signifying whether the provided evidence can provide sufficient clues to verify the claim. The claim-evidence relevance probability is calculated:
\begin{equation}
   P(y_{e_p}|c,e_p)=\text{Softmax}_{y_{e_p}}(\text{Linear} (h_p)),
\end{equation}
where $h_p$ denotes the initial node representation of node $n_p$.

\section{Experimental Methodology}\label{experimental}
This section describes the datasets, evaluation metrics, baselines, and implementation details used in our experiments.

\subsection{Dataset}
In our experiments, we leverage the FEVER~\cite{thorne2018fever} dataset and SCIFACT~\cite{wadden-etal-2020-fact} dataset, which focuses on the general domain and the science-specific domain, respectively. They are publicly available collections for the fact verification task. 
FEVER employs Wikipedia as its trustworthy corpus for claim verification. It comprises 185,455 annotated claims classified into SUPPORTS, REFUTES, or NOT ENOUGH INFO categories.
SCIFACT comprises 1,409 annotated claims sourced from 5,183 scientific articles. These claims are categorized as SUPPORT, CONTRADICT, or NOT ENOUGH INFO.
The data statistics of the FEVER dataset and the SCIFACT dataset are shown in TABLE~\ref{tab:dataset}. 

\begin{table}[t]
\center
\caption{Data statistics of FEVER Dataset and SCIFACT dataset.}
\begin{tabular}{l  |  l|  r r r}
\hline  \textbf{Dataset} & \textbf{Split} & \textbf{SUPPORT} & \textbf{REFUTE}& \textbf{NEI}\\ \hline

\hline
\multirow{3}{*}{FEVER} &Train    & 80,035 & 29,775&35,639 \\ 
& Dev &  6,666 & 6,666 & 6,666\\ 
&Test &  6,666 & 6,666 & 6,666  \\
\hline
\multirow{3}{*}{SCIFACT}& Train    & 332 &173 & 304 \\ 
&Dev &124& 64& 112\\
&Test& 100 &100 &100\\

\hline
\end{tabular}

\label{tab:dataset}
\end{table}

Similar to the previous fact verification systems~\cite{chen2022loren,DBLP:conf/coling/ParkLJKKN22}, our experiment employs the identical experimental configuration as KGAT~\cite{DBLP:conf/acl/LiuXSL20} on the FEVER dataset. We directly employ the document retrieval model and evidence retrieval model from KGAT. During the TEST stage, we need to submit our results to the designated website\footnote{\url{https://codalab.lisn.upsaclay.fr/competitions/7308}} for blind evaluation. We utilize the same experimental configuration as SCIKGAT~\cite{liu2020adapting} on the SCIFACT dataset. we also submit our results to the designated website\footnote{\url{https://leaderboard.allenai.org/scifact/submissions/public}} for blind evaluation. The experimental data can be obtained via GitHub\footnote{\url{https://github.com/thunlp/KernelGAT}}.

\subsection{Evaluation Metrics}
\textbf{FEVER:}
The same as the previous research~\cite{lee2021towards, DBLP:conf/acl/LiuXSL20}, CO-GAT uses the official evaluation metrics\footnote{\url{https://github.com/sheffieldnlp/fever-scorer}} FEVER score (FEVER) and Label Accuracy (ACC) to estimate the label prediction performance of model on the FEVER dataset. FEVER score is the primary evaluation metric of the CO-GAT model on the FEVER dataset, which offers a more comprehensive reflection of the inference capability. 

\textbf{SCIFACT:}
Precision, Recall, and F1 score are employed to assess the performance of the CO-GAT model on the SCIFACT dataset. It is evaluated at abstract and sentence levels.

\subsection{Baselines}
In our experiments, we compare the CO-GAT model with some fact verification work that focuses on graph reasoning or achieves fine performance on the FEVER dataset and SCIFACT dataset.

Athene~\cite{hanselowski2018ukp}, UNC NLP~\cite{nie2019combining} and UCL MRG~\cite{yoneda2018ucl} are three top models in FEVER 1.0 shared task~\cite{thorne2018fact}. They are based on the NLI model to establish factual reasoning models. In addition, they use attention mechanisms to focus on the semantic information of the claim and evidence.

The prevalence of pre-trained language models and graph neural network architectures has led to their integration into fact verification tasks, resulting in notable performance improvements. GEAR~\cite{zhou2019gear} represents an early attempt to establish a reasoning model for fact verification based on the graph neural network. It utilizes the graph-based evidence reasoning network to aggregate evidence information. The BERT model~\cite{devlin2019bert} is used to encode the claim and evidence sentence pairs. 
Subsequently, KGAT~\cite{DBLP:conf/acl/LiuXSL20} proposes a fine-grained graph inference model with a kernel-based graph attention network. It can utilize BERT~\cite{devlin2019bert}, RoBERTa~\cite{liu2019roberta}, and CorefRoBERTa~\cite{ye2020coreferential} models to encode graph nodes.
KGAT employs both the token-level and sentence-level kernel attention mechanisms to capture fine-grained evidence information. EvidenceNet~\cite{Chen-evidencenet} employs the gating mechanism to filter redundant evidence information and utilizes the evidence fusion network to capture global contextual information from different levels of evidence. ICMI~\cite{wang-etal-2022-imci} uses the graph neural network to integrate multi-view contextual information for fact verification, which includes the intra-document context of evidence sentences from the same document and the inter-document context of sentences from other documents.
DREAM~\cite{zhong-etal-2020-reasoning} is different from previous work, which establishes a semantic graph by employing a semantic role labeler to decompose claims and evidence sentences. Crucially, we build a graph-based baseline model MHA-GAT as our main baseline. It only uses the multi-head attention mechanism~\cite{vaswani2017attention} to establish a graph attention model to aggregate evidence information. 

Different from the graph-based inference models, MLA~\cite{DBLP:conf/acl/KruengkraiYW21} proposes a sequence inference model. It utilizes both token-level and sentence-level self-attentions to capture information and incorporates the static positional encoding mechanism into the input of the multi-head attention mechanism.
LOREN~\cite{chen2022loren} propose an interpretable fact verification at the phrase level. The veracity of the phrases serves as an explanation and is aggregated into the final verdict according to the logical rules.

\subsection{Implementation Details} 
The rest of this section describes the implementation details of the CO-GAT model on the FEVER dataset and SCIFACT dataset.

\begin{table}[t]
\begin{center}
\caption{\label{tab:retrieval}The Performance of Evidence Sentence Retrieval. The Precision, Recall, and F1 are calculated by official evaluation~\cite{thorne2018fever}.
}
\begin{tabular}{l|l|ccc}
\hline  & \textbf{Model }& \textbf{Prec@5} & \textbf{Rec@5} & \textbf{F1@5}\\ \hline
\multirow{5}{*}{Dev} &UNC NLP&36.49 &86.79 &51.38\\
&GEAR&40.60 &86.36& 55.23\\
&DREAM&26.67& 87.64 &40.90\\
&MLA&25.63& 88.64 &39.76\\
&ICMI&25.74 &92.86 &40.30 \\
&CO-GAT& 27.29 &94.37 &42.34 \\ 

\hline
\multirow{2}{*}{Test}
&MLA&25.33 &87.58 &39.29\\
&CO-GAT&25.21& 87.47 &39.14\\
\hline
\end{tabular}

\end{center}
\end{table}

\begin{table*}[th]
\center
\caption{Overall Performance. Different fact verification models have been evaluated on the FEVER dataset and SCIFACT dataset to measure their performance. The best evaluation results are highlighted in bold, and the results of CO-GAT are underlined. $\star$ represents the model that has not released the code.}
\resizebox{\textwidth}{!}{
\begin{tabular}{ l | l |c  c | c  c |c  c c |c  c c|c  c c| c  c c}

\hline \multirow{4}{*}{\textbf{}} & \multirow{4}{*}{\textbf{Model}} &\multicolumn{4}{c|}{\textbf{FEVER}} &  \multicolumn{12}{c}{\textbf{SCIFACT}} \\\cline{3-18}
&& \multicolumn{2}{c|}{\textbf{Dev}} &  \multicolumn{2}{c|}{\textbf{Test}}&\multicolumn{6}{c|}{\textbf{Dev}} &  \multicolumn{6}{c}{\textbf{Test}}\\ \cline{3-18}
&& \multirow{2}{*}{\textbf{ACC}}& \multirow{2}{*}{\textbf{FEVER}}& \multirow{2}{*}{\textbf{ACC}}& \multirow{2}{*}{\textbf{FEVER}}&\multicolumn{3}{c|}{\textbf{SentenceLevel}} &  \multicolumn{3}{c|}{\textbf{AbstractLevel}}&\multicolumn{3}{c|}{\textbf{SentenceLevel}} &  \multicolumn{3}{c}{\textbf{AbstractLevel}}\\ 
 && &  &  & & \textbf{Prec} & \textbf{Rec} & \textbf{F1} &\textbf{Prec} & \textbf{Rec} & \textbf{F1} & \textbf{Prec} & \textbf{Rec} & \textbf{F1} & \textbf{Prec} & \textbf{Rec} & \textbf{F1}\\ 
 \hline
& Athene~\cite{hanselowski2018ukp}& 68.49 &64.74& 65.46& 61.58&- &-&- &-&- &-&- &-&- &-&- &-\\
&UCL MRG~\cite{yoneda2018ucl} & 69.66 &65.41& 67.62 &62.50&- &-&- &-&- &-&- &-&- &-&- &-\\
&UNC NLP~\cite{nie2019combining} &69.72&66.49&68.21&64.21&- &-&- &-&- &-&- &-&- &-&- &-\\
\hline
\multirow{8}{*}{Base}&GEAR (BERT)~\cite{zhou2019gear}       &74.84	&70.69	&71.60	&67.10&55.32 	&14.21 	&22.61 &	64.29 	&17.22 &	27.17 &53.77&15.41&23.95&67.21&18.47&28.98
\\
&KGAT (BERT)~\cite{DBLP:conf/acl/LiuXSL20}		&78.02	&75.88		&72.81	&69.40&54.50 	&29.78 &	38.52 	&60.71 	&32.54 	&42.37&53.42&33.78&41.39&65.57&36.04&46.51  
\\
&MLA (RoBERTa)~\cite{DBLP:conf/acl/KruengkraiYW21}&77.54 &74.41 &-&-&59.07 	&38.25 	&46.43 	&62.96 &	40.67 &	49.42 &52.35&39.19&44.82&64.00&43.24&51.61
\\
&$\text{DREAM (XLNet)}^\star$~\cite{zhong-etal-2020-reasoning} & \textbf{79.16} &- &\textbf{76.85}& 70.60&- &-&- &-&- &-&- &-&- &-&- &-\\
&$\text{EvidenceNet (BERT)}^\star$~\cite{Chen-evidencenet} &78.53 &75.65 &73.31 &70.85&- &-&- &-&- &-&- &-&- &-&- &-\\
&MHA-GAT (RoBERTa)&74.67	&72.51	&70.61	&66.88 &61.20 	&30.60 &	40.80 	&62.39 &	32.54 	&42.77 &56.28	&32.70&	41.37	&63.79	&33.33	&43.79\\
&CO-GAT (RoBERTa) &\underline{77.74}	&\underline{75.86}	&\underline{73.27}	&\underline{70.41}&\underline{62.07} 	&\underline{34.43} 	&\underline{44.29} &	\underline{66.95}& \underline{37.80}	 	&\underline{48.32} &\underline{55.79}	&\underline{35.14}	&\underline{43.12}	&\underline{66.41}&\underline{39.19}		&\underline{49.29}\\
&MHA-GAT (ELECTRA) &78.14	&76.05&73.67	&70.28 &58.44 	&36.89 &	45.23 &	65.19 &	42.11 	&51.16 &55.30	&39.46	&46.06	&\textbf{67.13}	&43.24	&52.60
\\
&CO-GAT (ELECTRA)	&\underline{78.84	}&\underline{\textbf{76.77}}	&\underline{74.56}	&\underline{\textbf{71.43}}&\underline{\textbf{63.39}} 	&\underline{\textbf{38.80}} &	\underline{\textbf{48.14}} 	&\underline{\textbf{72.00}} 	&\underline{\textbf{43.06}} 	&\underline{\textbf{53.89}} &\underline{\textbf{58.08}}	&\underline{\textbf{40.81}}&\underline{47.94}		&\underline{67.11}	&\underline{\textbf{45.05}}	&\underline{\textbf{53.91}}\\

\hline
\multirow{10}{*}{Large}

&KGAT (BERT)~\cite{DBLP:conf/acl/LiuXSL20}&77.91	&75.86	&73.61	&70.24&51.33 &	31.69 &	39.19 	&58.06 &	34.45 	&43.24&46.86&34.32&39.63&58.87&37.39&45.73\\

&LOREN (BERT)~\cite{chen2022loren}		&78.44	&76.21&	74.43	&70.71&- &-&- &-&- &-&- &-&- &-&- &-\\

&KGAT (RoBERTa)~\cite{DBLP:conf/acl/LiuXSL20}&78.29	&76.11&	74.07	&70.38 &69.70 	&37.70 	&48.94 	&79.28 	&42.11 	&55.00&58.42&44.05&50.23&72.67&49.10&58.60\\

&LOREN (RoBERTa)~\cite{chen2022loren}		&81.14	&78.83	&76.42	&72.93&- &-&- &-&- &-&- &-&- &-&- &-\\
&$\text{EvidenceNet (RoBERTa)}^\star$~\cite{Chen-evidencenet} &81.46 &78.29 &76.95& 73.78&- &-&- &-&- &-&- &-&- &-&- &-\\
&$\text{IMCI (RoBERTa)}^\star$~\cite{wang-etal-2022-imci} &- &-&77.25 &\textbf{73.96}&- &-&- &-&- &-&- &-&- &-&- &-\\
&MLA (RoBERTa)~\cite{DBLP:conf/acl/KruengkraiYW21}		&79.31&	75.96&77.05	&73.72&\textbf{73.54}&44.81&55.69 &	\textbf{80.62} 	&49.76 	&61.54 &59.64&45.14&\textbf{51.38}&\textbf{74.32}&49.55&59.46\\
&MHA-GAT (RoBERTa) &80.21	&77.82	&76.24	&72.17&66.96 	&41.53 &	51.26 &	73.48& 	46.41 &	56.89 	&\textbf{59.78}	&44.59&	51.08&	73.97	&48.65	&58.70\\
&CO-GAT (RoBERTa)	&\underline{81.56}	&\underline{79.21}&	\underline{76.95}	&\underline{73.48}&\underline{66.67 }	&\underline{46.45} &	\underline{54.75} &	\underline{75.86} 	&\underline{52.63} &\underline{62.15}	&\underline{55.52}		&\underline{44.86}	&\underline{49.63}	&\underline{70.00}	&\underline{50.45}	&\underline{58.64} \\
&MHA-GAT (ELECTRA)&80.99	&78.47	&76.95&	72.56&71.01 	&46.17& 	55.96 	&78.99 	&52.15 &	62.82 &	51.64	&38.38	&44.03	&65.33&	44.14	&52.69 \\
&CO-GAT (ELECTRA)	&\underline{\textbf{81.65}}	&\underline{\textbf{79.32}}	&\underline{\textbf{77.27}}	&\underline{73.59}&\underline{71.49}	 &	\underline{\textbf{48.63}} &	\underline{\textbf{57.89}} 	&\underline{79.58 	}&\underline{\textbf{54.07}}&\underline{\textbf{64.39}} 	&\underline{55.31}	&\underline{\textbf{47.84}}	&\underline{51.30}	&\underline{69.64}	&\underline{\textbf{52.70}}	&\underline{\textbf{60.00}} \\
\hline



\end{tabular}
}

\label{tab: overall}
\end{table*}

\subsubsection{FEVER}
CO-GAT implements the document retrieval model following the previous fact verification systems~\cite{hanselowski2018ukp, DBLP:conf/ecir/SoleimaniMW20,zhou2019gear}. Firstly, we utilize the constituency parser of AllenNLP~\cite{gardner2018allennlp} to extract phrases from claims. Subsequently, these phrases serve as queries and use the online MediaWiki API\footnote{\url{https://www.mediawiki.org/wiki/API: Main_page}} to retrieve the wiki pages, whose titles exhibit the most significant overlap with the queries. The seven highest-ranked results of each query are selected as the candidate set. Finally, we filter out the candidate set by removing the pages whose title is longer than the phrase mentioned and does not overlap with the rest of the claim~\cite{hanselowski2018ukp}.

In our experiments, CO-GAT uses the BERT-based sentence retrieval model, which is proposed by the KGAT model~\cite{DBLP:conf/acl/LiuXSL20}. Firstly, it leverages the last hidden state of the ``[CLS]'' token to represent the claim and evidence pair. Then, the rank model projects the ``[CLS]'' representation to a rank score. Finally, we use the pairwise loss for training the retrieval model. The sentence retrieval performances of the baseline models are shown in TABLE~\ref{tab:retrieval}.

For the claim verification step, we inherit huggingface's implementation\footnote{https://github.com/huggingface/transformers} of the pre-trained language models to encode the claim and evidence pairs. When the claim-evidence pair exceeds the maximum length of 256, we employ the tail truncation method. During training, we set the epoch to 10 and the evaluation step to 1,000. Additionally, we employ the early stopping training method and set patience to 5. For the base model, we set the batch size to 16 and the accumulation step to 1. For the large model, we set the batch size to 8 and the accumulation step to 2.
The Adam optimizer is employed for training CO-GAT. For the base model, the learning rate is set to 5e-5.
For the large model, we implement a warm-up training strategy. Initially, we set the learning rate to 5e-5 and only train the attention layer. Subsequently, we load the checkpoint trained from the first stage and set the learning rate to 2e-6 to fine-tune all layers. The number of the attention head equals the dimension of the pre-trained language model divided by 64.

\subsubsection{SCIFACT}
For the abstract retrieval model, we retrieve the top-100 abstracts using the TF-IDF, the same as the previous work~\cite{wadden-etal-2020-fact}. Subsequently, it utilizes the hidden state of the "[CLS]" token as the representation of the claim and abstract pairs. Finally, we introduce a ranking model to select the top-3 abstracts. We kept the same setting as SCIKGAT~\cite{liu2020adapting}, which set the max length to 256, learning rate to 2e-5, batch size to 8, and accumulate steps to 4.

The rationale selection model aims to select the relevant sentence from the retrieved abstract for claim verification. Specifically, we use the pre-trained model to encode the claim and evidence pairs and use the last hidden state of the ``[CLS]'' token to predict the relevance label. We keep the same setting as the previous work~\cite{liu2020adapting, wadden-etal-2020-fact}.

For the claim verification model, CO-GAT used the same settings implemented on the FEVER dataset. The pre-trained language models also inherit huggingface's PyTorch implementation to encode the claim and evidence pairs. The difference is that for evidence selection, we exclusively selected evidence sentences without incorporating titles from wiki documents.

\section{Evaluation Result}\label{result}
In this section, we describe the experiments, which are conducted on the FEVER dataset and SCIFACT dataset, to evaluate the performance of CO-GAT on the fact verification task.
Firstly, we present the overall performance of the CO-GAT model on the fact verification task. Then we conduct ablation studies and explore the tendency of the prediction label. 
We also study the effectiveness of the node masking mechanism and the mode of the CO-GAT attention mechanism. Finally, we present case studies.

\subsection{Overall Performance}
The fact verification performance of CO-GAT on the FEVER dataset and SCIFACT dataset are compared with several baseline models. The results are shown in TABLE~\ref{tab: overall}.

Overall, CO-GAT confirms its effectiveness by achieving a 73.59\% FEVER score on the blind test set. Compared to these graph-based claim verification models~\cite{zhou2019gear,zhong-etal-2020-reasoning,DBLP:conf/acl/LiuXSL20}, our CO-GAT model achieves a 3.35\% improvement. It shows that our node masking mechanism is effective in enhancing the graph reasoning process and does not change the vanilla graph modeling architecture.
Compared with the MHA-GAT model, CO-GAT achieves a 1.03\% improvement by keeping the same graph reasoning module as the MHA-GAT model. It shows that the node masking mechanism can help filter out the noise information and conduct more calibrated node representations before feeding the node representations to the reasoning graph. Furthermore, CO-GAT also demonstrates its effectiveness by keeping the vanilla GAT architecture for graph reasoning and conducting competitive performance with EvidenceNET and IMCI models. CO-GAT avoids conducting repetitive encoding~\cite{Chen-evidencenet} and building an additional reasoning graph in the fact verification model~\cite{wang-etal-2022-imci}. 

Besides the FEVER dataset, we also conduct experiments on the SCIFACT dataset, which focuses on scientific claim verification. CO-GAT also outperforms baseline models on the SCIFACT dataset, which also confirms its effectiveness. The better fact verification results show the generalization ability of CO-GAT by broadening its effectiveness to the science-specific domain. 

\begin{table}[t]
\center
\caption{Ablation Study. $\dagger$ $\ddag $ $\S$ indicate statistically significant improvements over $\text{CO-GAT (hard CO-SCO)}^\dagger$, $\text{CO-GAT w/o Node Mask}^\ddag $ and ${\text{CO-GAT w/o }{L_{evi}} ^  \S }$.}
\resizebox{\linewidth}{!}{
\begin{tabular}{ l|l |l  l |l l }
\hline \multirow{2}{*}{\textbf{}} & \multirow{2}{*}{\textbf{Model}} & \multicolumn{2}{c|}{\textbf{Dev}}& \multicolumn{2}{c}{\textbf{Test}}\\ \cline{3-6}
 & &\textbf{ACC} & \textbf{FEVER} &\textbf{ACC} & \textbf{FEVER}\\ \hline



\multirow{4}{*}{ELECTRA}		
&$\text{CO-GAT (hard CO-SCO)}^\dagger$&81.12 &78.53  &77.10 &72.70 \\
&$\text{CO-GAT}$		&$\text{81.65}^ {\dagger \ddag }$&	\text{79.32}$\text{}^ {\dagger \ddag}$&77.27&73.59\\

&${\text{CO-GAT w/o }{L_{evi}} ^  \S }$& $\text{81.65}^ {\dagger \ddag}$ &$\text{79.14}^ {\dagger \ddag}$&77.59 &73.42 \\
&$\text{CO-GAT w/o Node Mask}^\ddag $&80.99	&78.47&76.95&72.56	\\
\hline
\multirow{3}{*}{T5-Large}
&Concat&79.61&77.64&75.61&72.43\\
&MHA-GAT&80.38&78.01&75.72&72.05\\
&CO-GAT&80.45&78.26&76.01&72.60\\
\hline
\multirow{3}{*}{GPT2}
&Concat&76.30&74.08&72.06&68.09\\
&MHA-GAT&76.28&74.12&71.34&67.73\\
&CO-GAT &76.57&74.54&71.75&68.35\\
\hline
\end{tabular}
}

\label{tab:ablation}
\end{table}

\subsection{Ablation Study}\label{sec:ablation}
This section investigates the effectiveness of different modules and backbone models of the CO-GAT model through ablation studies.

\subsubsection{The Effect of Using Node Masking Mechanism}
In this experiment, we conduct two models, the \text{CO-GAT} (hard CO-SCO) model and the CO-GAT w/o Node Mask model to explore the efficacy of the node masking mechanism used in the CO-GAT model.

Different from the confidence score used in the CO-GAT model, which sets the probability between 0 and 1, the CO-GAT (hard CO-SCO) model sets the node mask probability to 0 or 1. If the confidence score is 0, the model will cover up the entire piece of evidence with the blank node, otherwise, the evidence nodes will maintain their initial representation.

As shown in TABLE~\ref{tab:ablation}, the CO-GAT model achieves 0.85\% improvements of the FEVER score, compared with the CO-GAT model w/o Node Mask model (MHA-GAT). This demonstrates the effectiveness of utilizing the node masking mechanism, which avoids the propagation of noise information in the graph reasoning. The CO-GAT model improves the fact verification performance by 0.79\% compared with the CO-GAT (hard CO-SCO) model. Due to the inevitable occurrence of incorrect hard-label predictions, the claim-related node may be completely erased. Therefore, a soft score that partly erases the node information can achieve more precision in the claim verification model. In the following section (Sec.~\ref{modedl: sensitive}), we further explore the sensitivity of the confidential score.

\begin{figure}[t]
\centering
\subfloat[The Relationship between the NEI Prediction Probability and Cross Entropy Scores.]{\includegraphics[width=1.65in]{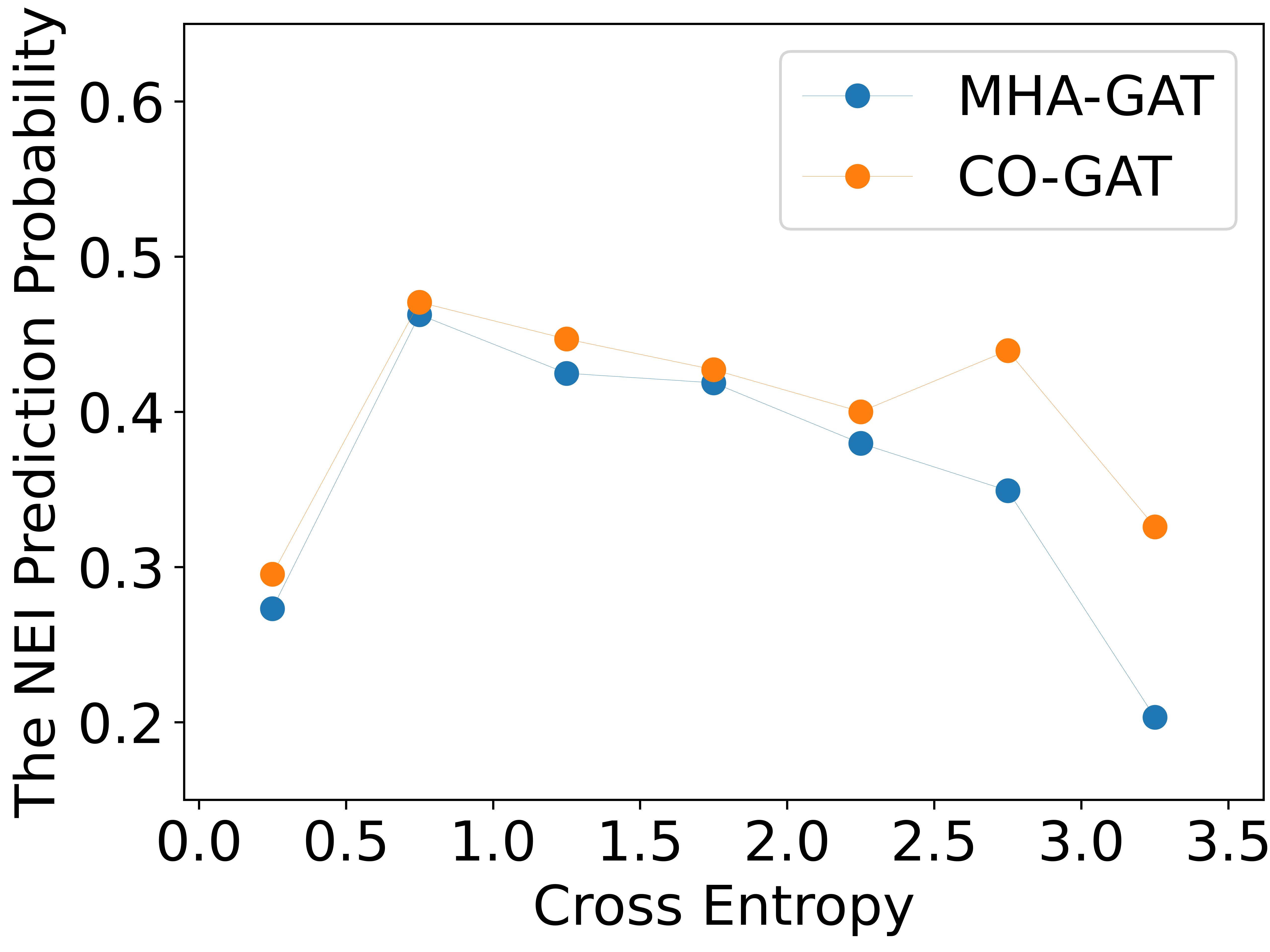}
\label{fig：nei_all}}
\hfill
\subfloat[NEI Prediction Ratio of Different Claim Labels.]{\includegraphics[width=1.65in]{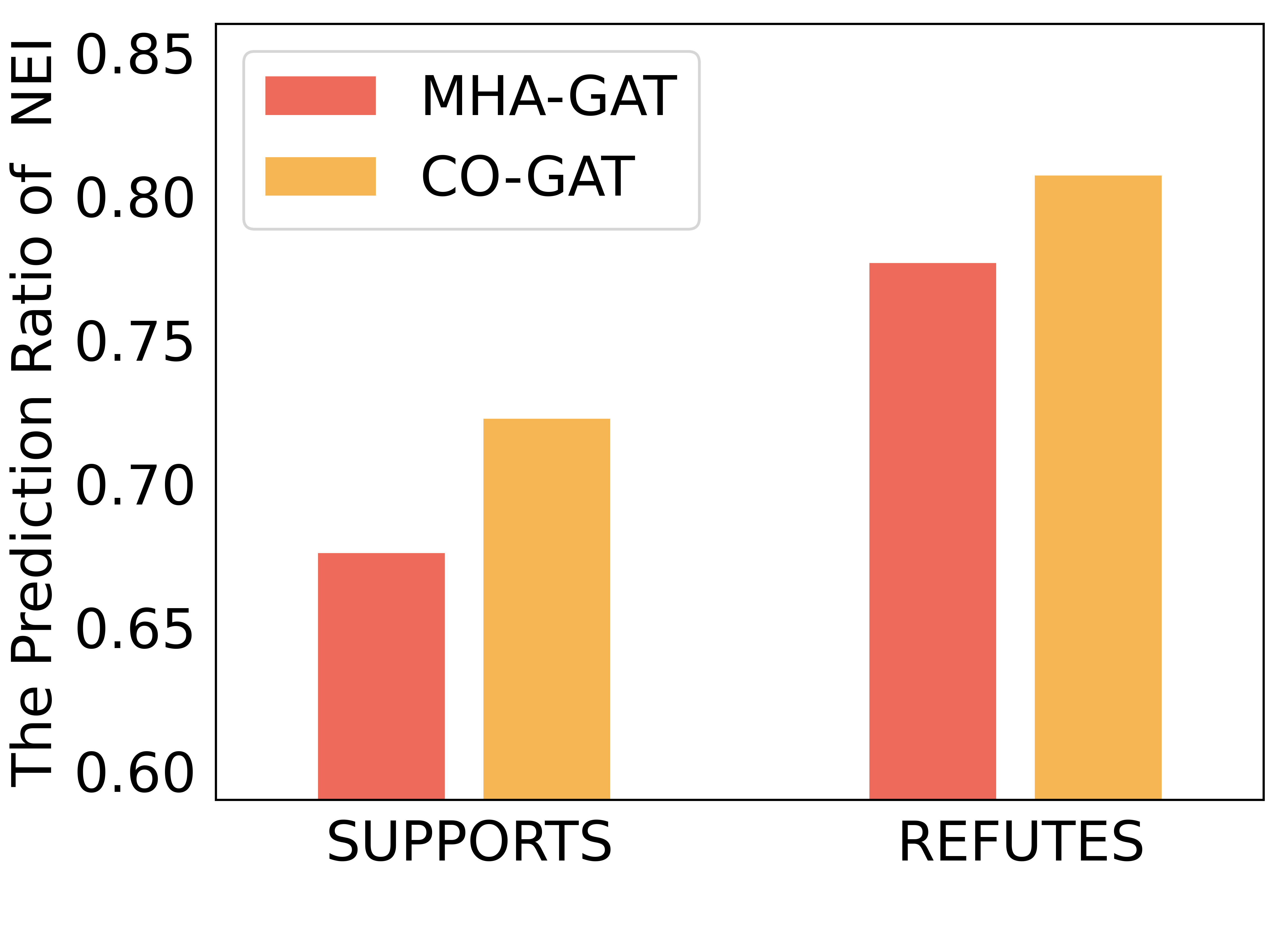}
\label{fig:nei_false}}
\hfill
\caption{Tendency and Distribution of NEI Label Prediction. Fig.~\ref{fig：nei_all} represents the prediction probability of NEI changing with cross-entropy. The increase of cross entropy indicates the higher uncertainty of the model~\cite{DBLP:conf/emnlp/WangLWLS19,DBLP:conf/icml/OttAGR18}. Fig.~\ref{fig:nei_false} represents the NEI distribution in the cases that make a mistake prediction.}
\label{fig:neipro}
\end{figure} 
\subsubsection{The Effectiveness of Multi-Task Modeling Methods}
 In this experiment, we change the training strategy to evaluate the effectiveness of the multi-task training and compare CO-GAT with the CO-GAT w/o $L_{evi}$ model. 
 
 During training, the CO-GAT w/o $L_{evi}$ model only uses the claim verification loss ($L_{fact}$), while the CO-GAT model uses both the claim verification loss ($L_{fact}$) and the node prediction loss ($L_{evi}$). Compared with the CO-GAT w/o $L_{evi}$ model, the CO-GAT model achieves a 0.18\% improvement of 0.18\%, showing that the claim-evidence relevance signals can benefit the node masking mechanism by learning a more accurate confidence score (CO-SCO). 

\subsubsection{The Effectiveness of Different Backbone Models}
In this experiment, we investigate the influence of the backbone model. We employ the encoder-base models (ELECTRA and RoBERTa), decoder-based model (GPT2), and encoder-decoder based model (T5) to assess the performance of CO-GAT. The concat model combines claims and all evidence pieces. Following sentence-t5~\cite{ni-etal-2022-sentence}, the CO-GAT (T5) model utilizes T5~\cite{raffel2020exploring} to encode the claim-evidence pairs and get the node representation by using the encoded representation of the first input token of the T5 decoder module. The GPT2 model feeds claim-evidence pairs and uses the end token of the text to obtain the node representation. When employing the GPT2 model and T5 model as the backbone model, the CO-GAT model also exhibits an improving fact verification performance compared with MHA-GAT, which further confirms the effectiveness of CO-GAT and demonstrates its generalization ability.

\begin{figure*}[t]
\centering
\subfloat[NEI Probability.]{\includegraphics[width=0.33\linewidth]{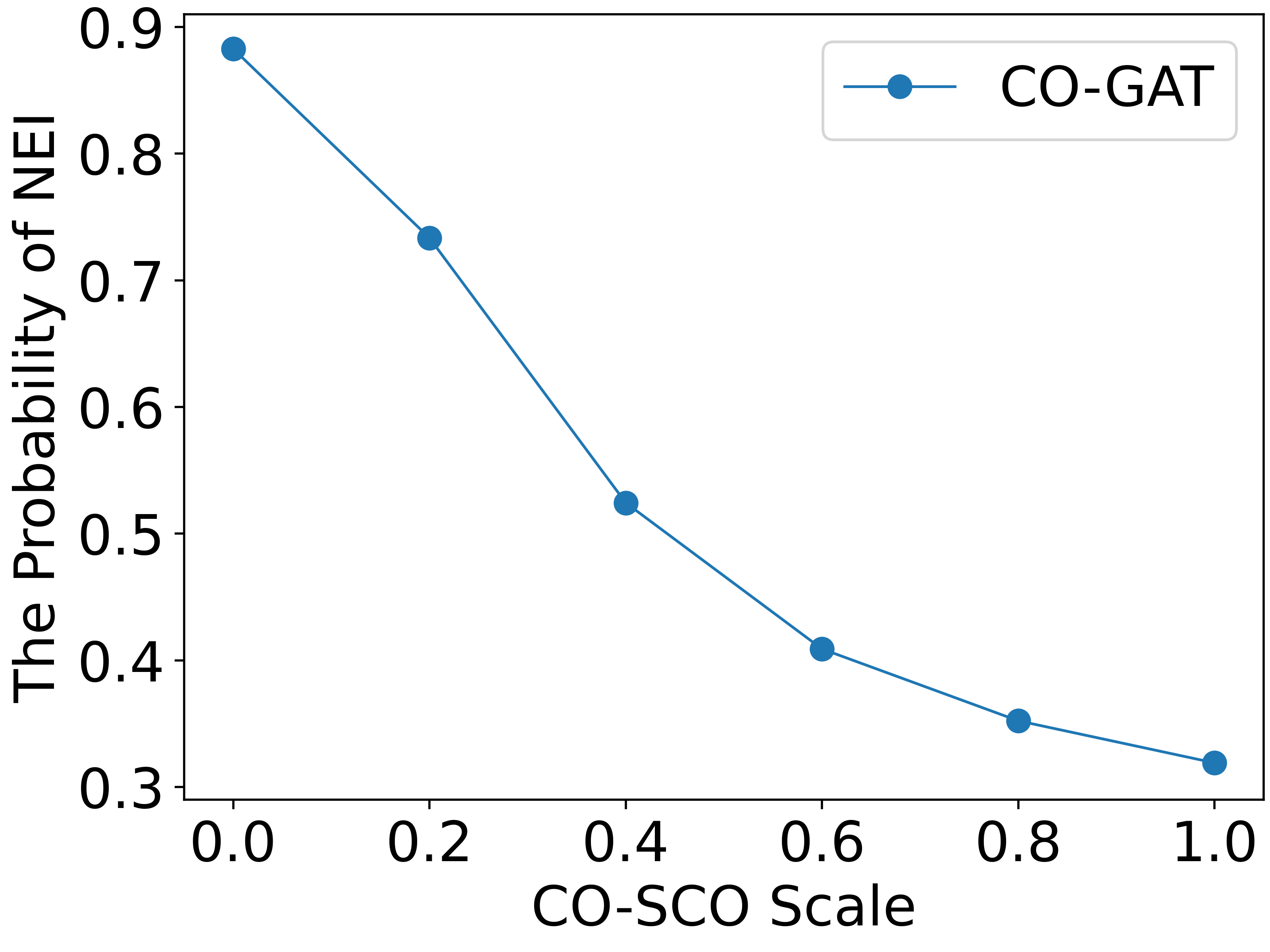}
\label{fig_second_case}}
\subfloat[Label Accuracy.]{\includegraphics[width=0.33\linewidth]{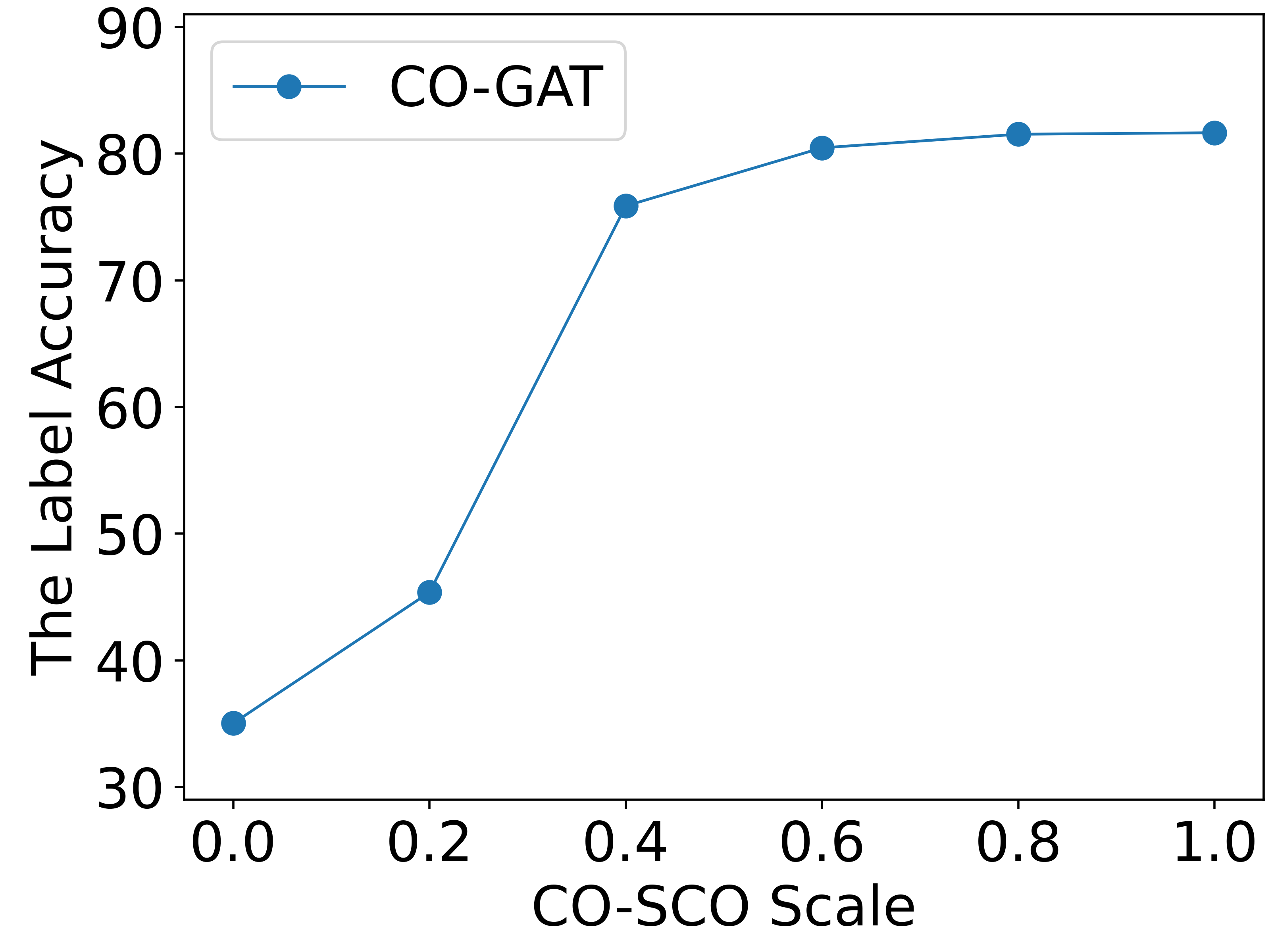}
\label{fig_first_case}}
\subfloat[Attention Entropy.]{\includegraphics[width=0.33\linewidth]{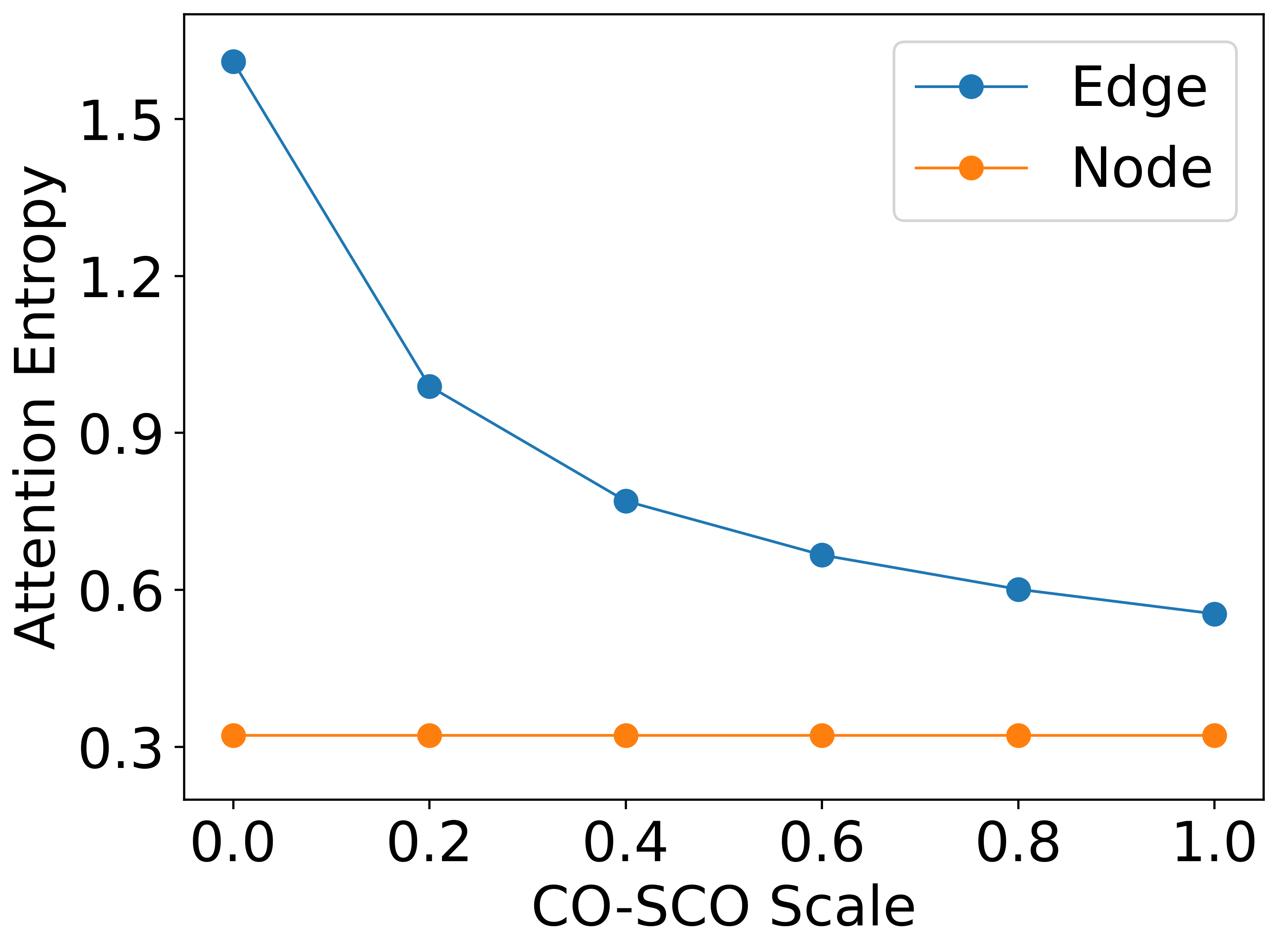}
\label{fig_third_case}}

\caption{The Effectiveness of Confidence Score in Controlling Information Flows from Encoded Node Representations to Graph Reasoning. Fig.~\ref{fig_second_case} describes the change in the probability of predicting the label as NEI. Fig.~\ref{fig_first_case} describes the tendency of label accuracy. Fig.~\ref{fig_third_case} describes the tendency of edge attention entropy and node attention entropy. We scale the CO-SCO by multiplying the original confidence score using a coefficient $\alpha \in [0.0, 1.0]$.}
\label{fig:scale}
\end{figure*}

\subsection{Label Prediction Behaviors of CO-GAT}
In Fig.~\ref{fig:neipro}, we conduct two experiments: the initial experiment investigates the correlation between NEI label prediction probability and the cross entropy score; the subsequent one analyses the NEI prediction ratio of the cases that belong to SUPPORT and REFUTE.

As shown in Fig.~\ref{fig：nei_all}, when the cross entropy is low, both MHA-GAT and CO-GAT exhibit a similar probability distribution in predicting the claim label as NEI. However, different from the MHA-GAT model, the CO-GAT model conducts a notably higher probability of predicting the claim label as NEI when the cross entropy score increases, which indicates that the CO-GAT model prefers to predict the claim label as NEI when the cross entropy is higher. Such a phenomenon demonstrates that the reasoning ability of CO-GAT relies more on sufficient information and calibrates the uncertain predictions to the NEI, making CO-GAT conduct more reliable predictions.

Furthermore, we analyze the NEI prediction ratios of the cases that belong to the SUPPORT and REFUTE classes. As shown in Fig.~\ref{fig:nei_false}, the results indicate that, in comparison to the MHA-GAT model, CO-GAT exhibits a higher tendency to predict the claim label as NEI. This further confirms the calibration behavior of our CO-GAT model, which predicts some confusing cases as the NEI label.

\subsection{The Effectiveness of Confidence Score in Controlling the Information Flows that from Node Representations to the Graph Reasoning Module}\label{modedl: sensitive}
In Fig.~\ref{fig:scale}, we scale the confidence score to explore the effectiveness of the confidence score in controlling the semantic information flow from the node representations to the graph reasoning module. 

As shown in Fig.~\ref{fig_second_case}, we scale the CO-SCO by multiplying it with a coefficient $\alpha$, which changes from 1.0 to 0.0 results with the interval of 0.2. As the confidence score decreases, more evidence information flows from blank nodes into the graph nodes. In this case, the reasoning model lacks sufficient evidence information to make a correct fact verification prediction, leading to the tendency of the NEI prediction of CO-GAT. Nonetheless, as shown in Fig.~\ref{fig_first_case}, the label accuracy exhibits only a slight decrease while the scaling factor varies from 1.0 to 0.4.
Despite inevitable errors in node confidence assessment, it ensures that CO-GAT can preserve useful semantic information of node representations and feed them to the reasoning graph, indicating its strong tolerance of the CO-SCO prediction in our node masking mechanism.

As shown in Fig.~\ref{fig_third_case}, we analyze the change of cross entropy scores by scaling the CO-SCO to control the information flow from the blank node to these initial node representations. The lower attention entropy score indicates a more concentrated attention mechanism. The evaluation results show that, as the scaling coefficient decreases from 1.0 to 0.0, the edge attention entropy gradually increases, which illustrates that the semantic information gradually flows from blank nodes to vanilla nodes. In this case, the node representations tend to become more homogeneous, making the edge attention mechanism unable to identify these useful nodes.
On the contrary, the node attention entropy remains relatively stable regardless of changes in the scaling coefficient. It demonstrates that the node representations become more homogenized after multi-layer graph attention encoding. This indicates that the edge attention mechanism is primarily responsible for node selection and the node attention mechanism does not pay attention to node selection. This further demonstrates the necessity of utilizing the node masking mechanism to erase noise information before the graph reasoning, preventing its impact on the learned representations of other nodes during graph reasoning.

\begin{figure*}[t]
\centering
\subfloat[Attention Entropy.]{
   \includegraphics[width=0.33\linewidth]{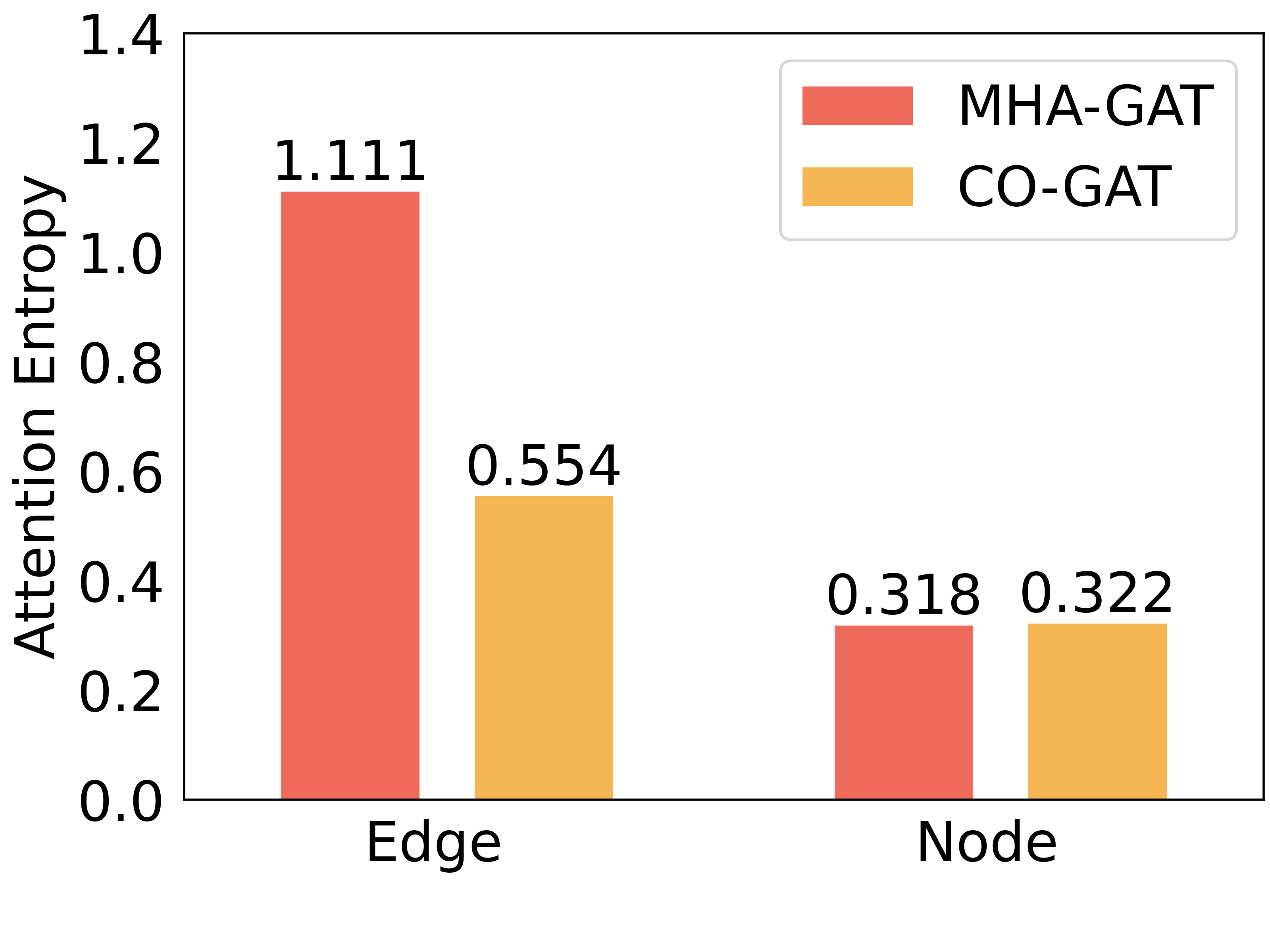}
    
\label{fig: 4a}}
\subfloat[Edge Attention Distribution.]{\includegraphics[width=0.33\linewidth]{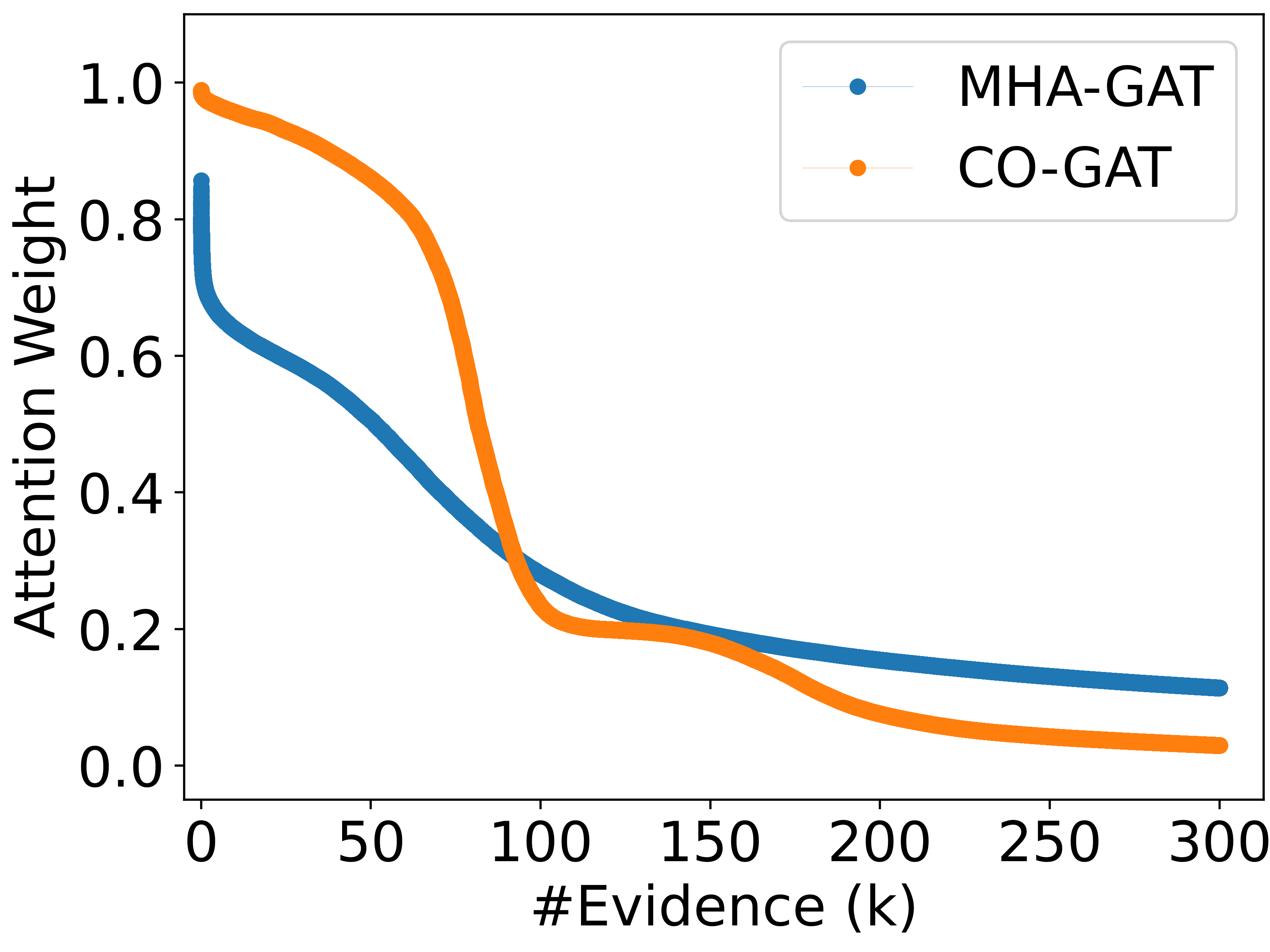}
\label{fig: 4b}}
\subfloat[Node Attention Distribution.]{\includegraphics[width=0.33\linewidth]{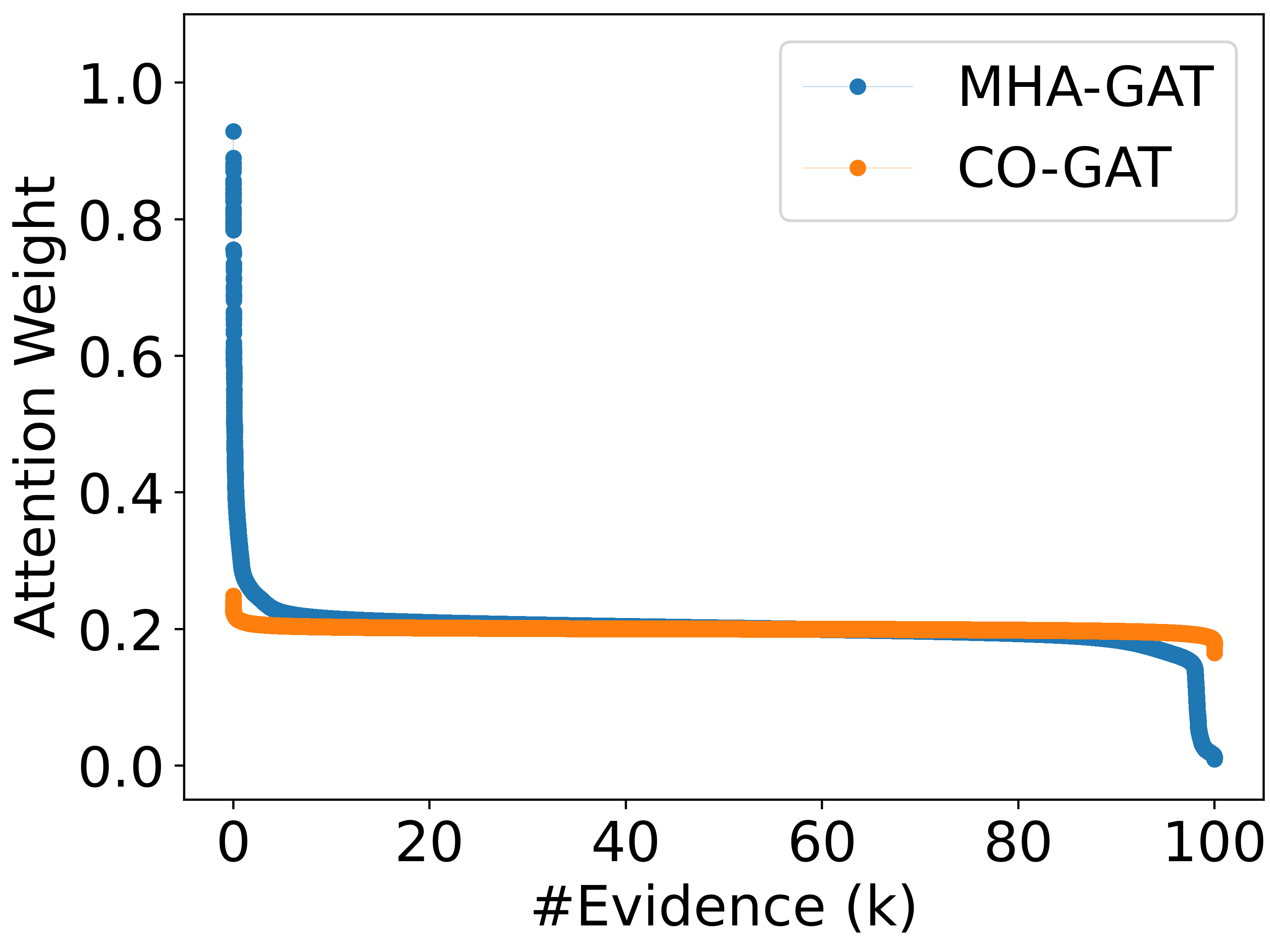}
\label{fig: 4c}}
\caption{The Behaviors of Edge Attention and Node Attention After Adding Node Masking Mechanism. Fig.~\ref{fig: 4a} shows the attention entropy of the edge and the node of the CO-GAT model and the MHA-GAT model, respectively. Fig.~\ref{fig: 4b} shows the edge attention weight distribution of the CO-GAT model and the MHA-GAT model. Fig.~\ref{fig: 4c} shows the node attention weight distribution of the CO-GAT model and the MHA-GAT model.}
\label{fig:att_evi}
\end{figure*} 
\subsection{The Analyses on Attention Mechanisms of CO-GAT}
In this experiment, we investigate the comparison of the attention mechanisms between the CO-GAT model and the MHA-GAT model. 

As shown in Fig.~\ref{fig: 4a}, the node attention entropy scores of MHA-GAT and CO-GAT are almost the same, which indicates that both of them have a similar node attention distribution. This shows that CO-GAT has little effect on node attention. On the contrary, the edge attention entropy of CO-GAT is notably smaller than that of MHA-GAT. It demonstrates that the edge attention mechanism of CO-GAT helps to conduct a more concentrated distribution while using edge attention to encode node representations. The main reason may lie in that the node masking mechanism has the ability to erase some unnecessary information from the initial node representations, make these node representations more distinguished, and finally conduct a more concentrated attention distribution.

Furthermore, we analyze the attention distribution of edge attention and node attention by sorting the attention weights.
As shown in Fig.~\ref{fig: 4b}, during the graph reasoning process, CO-GAT assigns more attention weights to some nodes, which demonstrates that our attention masking mechanism can expose the valuable node representations by alleviating the effect of noise node representations. As shown in Fig.~\ref{fig: 4c}, it illustrates that the node attention weights of the CO-GAT model are more stable, compared to the MHA-GAT model. These encoded node representations become more homogeneous after edge attention based encoding, which thrives on erasing the noise information from the initial node representations.

\begin{table*}[t]
\centering
\caption{Case Studies. We show three cases from the FEVER dataset. Node and Edge represent the weights of node and edge attention.}
\begin{tabular}{p{1.2\columnwidth}|p{0.1\columnwidth}<{\centering}|p{0.1\columnwidth}<{\centering}p{0.1\columnwidth}<{\centering}|p{0.10\columnwidth}<{\centering}p{0.1\columnwidth}<{\centering}}
\hline 
&\multirow{2}{*}{CO-SCO}&\multicolumn{2}{c}{CO-GAT}\vline &\multicolumn{2}{c}{MHA-GAT}\\
\cline{3-6}
&&Node&Edge&Node&Edge\\
\hline 
\multicolumn{4}{l}{\textbf{Case \#1}} 
\\
\hline 
\textbf{\textcolor{blue}{Claim:}} \textbf{\emph{\textcolor{red}{CHiPs}}} was \textbf{\emph{\textcolor{red}{created}}} in \textbf{\emph{\textcolor{red}{May 2017}}}.&&&&\\ 
\textcolor{blue}{\bm{$E_1$}:} [\textbf{\emph{\textcolor{red}{CHiPs}}} (film)] The film was released on March 24, 2017 by Warner Bros. &0.01 &0.2032 &0.0467 &0.1890&0.1148\\

\multicolumn{1}{m{1.21\columnwidth}}{\textcolor{blue}{\bm{$E_2$}:} [\textbf{\emph{\textcolor{red}{CHiPs}}} (film)] \textbf{\emph{\textcolor{red}{CHiPs}}} is a 2017 American action comedy buddy cop film written and directed by Dax Shepard, based on the 1977 1983 television series of the same name created by Rick Rosner.}\vline &0.02 &0.2030 &0.0467&0.1925&0.1078\\

\multicolumn{1}{m{1.21\columnwidth}}{\textcolor{blue}{\bm{$E_3$}:} [\textbf{\emph{\textcolor{red}{CHiPs}}}]\textbf{\emph{\textcolor{red}{CHiPs}}} is an American television drama series that originally aired on NBC from September 15, 1977, to May 1, 1983.}\vline&0.04 &0.2020 &0.0490&0.2133&0.3304\\

\multicolumn{1}{m{1.21\columnwidth}}{\textcolor{blue}{\bm{$E_4$}:} [\textbf{\emph{\textcolor{red}{CHiPs}}}] The series ran for 139 episodes over six seasons, plus one reunion TV movie from October 27, 1998.}\vline&0.12 &0.1995 &0.0562&0.2019&0.1449\\
\multicolumn{1}{m{1.21\columnwidth}}{\textcolor{blue}{\bm{$E_5$}:} [\textbf{\emph{\textcolor{red}{CHiPs}}} (film)] Principal photography began on October 21, 2015 in Los Angeles.}\vline&0.28 &0.1923 &0.8013&0.2033&0.3020\\
\hline
\multicolumn{4}{l}{Actual Label: NEI \hspace{8em}  CO-GAT: NEI \hspace{8em} MHA-GAT: REFUTES}\\
\hline
\multicolumn{1}{l}{\textbf{Case \#2} }         \\
\hline 
\textbf{\textcolor{blue}{Claim:}} 
\textbf{\emph{\textcolor{red}{Aphrodite}}} is the \textbf{\emph{\textcolor{red}{daughter}}} of a \textbf{\emph{\textcolor{red}{Titaness}}} in \textbf{\emph{\textcolor{red}{Homer's Iliad}}}.&&&&\\ 
\multicolumn{1}{m{1.21\columnwidth}}{\textcolor{blue}{\bm{$E_1$}:} \textcolor{violet}{(\textbf{Golden})}
[\textbf{\emph{\textcolor{red}{Aphrodite}}}] In \textbf{\emph{\textcolor{red}{Homer's Iliad}}}, however, she is the \textbf{\emph{\textcolor{red}{daughter}}} of Zeus and Dione.}\vline &0.84 	&0.2046 	&0.1725 	&0.2341 	&0.5542 \\
\multicolumn{1}{m{1.21\columnwidth}}{\textcolor{blue}{\bm{$E_2$}:} 
\textcolor{violet}{(\textbf{Golden})}
[\textbf{\emph{\textcolor{red}{Aphrodite}}}] 
Dione was an ancient Greek goddess, an oracular \textbf{\emph{\textcolor{red}{Titaness}}} primarily known from Book V of \textbf{\emph{\textcolor{red}{Homer's Iliad}}}, where she tends to the wounds suffered by her \textbf{\emph{\textcolor{red}{daughter}}} \textbf{\emph{\textcolor{red}{Aphrodite}}}.
} \vline&0.95 	&0.1982 	&0.7960 	&0.0258 	&0.0990\\
\multicolumn{1}{m{1.21\columnwidth}}{\textcolor{blue}{\bm{$E_3$}:} [\textbf{\emph{\textcolor{red}{Aphrodite}}}] She played a role in the Eros and Psyche legend, and was both lover and surrogate mother of Adonis.}\vline&0.01 	&0.1991 	&0.0105 	&0.3794 	&0.1155\\
\multicolumn{1}{m{1.21\columnwidth}}{\textcolor{blue}{\bm{$E_4$}:} [\textbf{\emph{\textcolor{red}{Aphrodite}}}] \textbf{\emph{\textcolor{red}{Aphrodite}}} is the Greek goddess of love, beauty, pleasure, and procreation.}\vline&0.01 	&0.1990 	&0.0105 	&0.0235 	&0.1048\\
\multicolumn{1}{m{1.21\columnwidth}}{\textcolor{blue}{\bm{$E_5$}:} [\textbf{\emph{\textcolor{red}{Aphrodite}}}] Many lesser beings were said to be children of \textbf{\emph{\textcolor{red}{Aphrodite}}}.}\vline&0.01 	&0.1991 	&0.0105 	&0.3373 	&0.1264\\

\hline
\multicolumn{4}{l}{Actual Label: SUPPORTS \hspace{5em}  CO-GAT: SUPPORTS \hspace{5em} MHA-GAT: REFUTES}\\

\hline 

\multicolumn{4}{l}{\textbf{Case \#3}}         \\
\hline
\textbf{\textcolor{blue}{Claim:}} \textbf{\emph{\textcolor{red}{The Nice Guys}}} is a \textbf{\emph{\textcolor{red}{romantic comedy}}}.&&&&\\
\multicolumn{1}{m{1.21\columnwidth}}{\textcolor{blue}{\bm{$E_1$}:}
\textcolor{violet}{(\textbf{Golden})}[\textbf{\emph{\textcolor{red}{The Nice Guys}}}] \textbf{\emph{\textcolor{red}{The Nice Guys}}} is a 2016 American neo noir \textbf{\emph{\textcolor{red}{action comedy}}} film directed by Shane Black and written by Black and Anthony Bagarozzi.} \vline&	0.29	&	0.1979 	&	0.8055 	&	0.1957 	&	0.5625 \\
\multicolumn{1}{m{1.21\columnwidth}}{\textcolor{blue}{\bm{$E_2$}:}[\textbf{\emph{\textcolor{red}{The Nice Guys}}}] The film stars Russell Crowe, Ryan Gosling, Angourie Rice, Matt Bomer, Margaret Qualley, Keith David and Kim Basinger.} \vline&	0.01	&	0.2005 	&	0.0485 	&	0.2057 	&	0.1071   \\
\multicolumn{1}{m{1.21\columnwidth}}{\textcolor{blue}{\bm{$E_3$}:}[\textbf{\emph{\textcolor{red}{The Nice Guys}}}] The Nice Guys premiered on May 15, 2016, at the 2016 Cannes Film Festival and was released by Warner Bros.} \vline&	0.01	&	0.2003 	&	0.0494 	&	0.1882 	&	0.0688  \\
\multicolumn{1}{m{1.21\columnwidth}}{\textcolor{blue}{\bm{$E_4$}:}[\textbf{\emph{\textcolor{red}{The Nice Guys}}}] Set in Los Angeles, 1977, the film focuses on a private eye LRB Gosling RRB and a tough enforcer LRB Crowe RRB who team up to investigate the disappearance of a teenage girl.} \vline&	0.01	&	0.2007 	&	0.0474 	&	0.2111 	&	0.1383   \\
\multicolumn{1}{m{1.21\columnwidth}}{\textcolor{blue}{\bm{$E_5$}:}[\textbf{\emph{\textcolor{red}{The Nice Guys}}}] It received positive reviews from critics but grossed just 57 million against its 50 million budget.} \vline&	0.01	&	0.2006 	&	0.0492 	&	0.1993 	&	0.1233  \\

\hline
\multicolumn{4}{l}{Actual Label: REFUTES \hspace{6em}  CO-GAT: NEI \hspace{8em} MHA-GAT: REFUTES}\\

\hline
\end{tabular}

\label{tab:case}
\end{table*}

\subsection{Case Studies}
TABLE~\ref{tab:case} shows three claim examples in the FEVER dataset. 

In the first case, ``CHIPs'' is ambiguous, which indicates two meanings: movies and TV dramas. However, this claim does not specify which was mentioned. The MHA-GAT model assigns approximately 0.3 edge attention weight to $E_3$ and $E_5$. Yet, $E_3$ states ``CHIPs aired from September 15, 1977, to May 1, 1983'', $E_5$ declares `` CHIPs began on October 21, 2015''. The timing of these two pieces of evidence is different from the ``2017'' in the claim. The presence of interference information ``1977'' and ``2015'' leads to incorrect prediction results in the MHA-GAT model, which predicts the claim label as ``REFUTES''. On the contrary, although the CO-GAT model assigns more edge attention weight to the $E_5$, it predicts the confidence score of $E_5$ to be 0.28. It erases the majority of noise information based on confidence scores and makes it more inclined to the blank node. Due to the lack of sufficient information remaining in the reasoning model, the CO-GAT model correctly predicts that the label of this claim is ``NEI''.

As shown in the second case, the MHA-GAT model pays more attention weight to the evidence $E_1$ and the other four pieces of evidence receive similar attention weights.
$E_1$ indicates that ``Aphrodite is the daughter of Dione, as mentioned in Homer's Iliad''. $E_2$ illustrates that ``Dione is an oracular Titaness primarily known from Book V of Homer’s Iliad''. $E_1$ and $E_2$ can jointly infer that the label of this claim should be ``SUPPORTS''. However, in the MHA-GAT model, the $E_2$ obtains a minimal node attention weight, which is only 0.02. It leads to the MHA-GAT model not obtaining sufficient effective information for claim verification. Meanwhile, $E_3$ and $E_5$ have a higher node attention weight compared to the $E_2$, which indicates ``Aphrodite'' is a ``mother''. This introduces additional noise semantic information, resulting in the MHA-GAT model predicting that the claim label is ``REFUTES''. In contrast, the $E_1$ and $E_2$ in the CO-GAT model have high confidence scores, which allows them to retain more useful information. Furthermore, $E_1$ and $E_2$ receive higher edge attention weights than the other evidence pieces, thus CO-GAT predicts the correct label ``SUPPORTS''.

In the third case, the MHA-GAT model and the CO-GAT model pay more attention to the golden evidence $E_1$.
The evidence $E_1$ states ``The Nice Guys is a 2016 American action comedy film''. Consequently, the MHA-GAT model predicts the claim label as ``REFUTES''. 
However, the CO-GAT model mistakenly evaluates the correlation between $E_1$ and the claim, leading to erasing the majority of its semantic information. Thus, the CO-GAT model makes a mistake prediction of the claim label as ``NEI''.

\section{Conclusion}\label{conclusion}
This paper proposes CO-GAT, a confidential graph reasoning model to verify the claim. It uses the blank node to erase the noise information of the initial node based on the confidential score. Our experiments show the model tends to predict the claim label as NEI when there is insufficient information to support or refute the claim. Our studies illustrate the node masking mechanism realized reasonable confidence score modeling, which controls the evidence information flow into the graph reasoning and prevents the propagation of noise signals in the graph reasoning. Our studies illustrate the node attention mechanism has a scattered distribution, which equally aggregates the node evidence representation. While, the edge attention mechanism plays the role of selecting useful nodes, which assigns more attention weight to the nodes that are related to the claim.

\section*{Acknowledgments}
This work is supported by the National Natural Science Foundation of China under Grant (No. U23B2019, No. 62072083, No. 62206042), the Joint Funds of Natural Science Foundation of Liaoning Province (No. 2023-MSBA-081) and the Fundamental Research Funds for the Central Universities under Grant (No. N2216017).

\bibliographystyle{IEEEtran}
\bibliography{re}

\begin{thebibliography}{10}
\providecommand{\url}[1]{#1}
\csname url@samestyle\endcsname
\providecommand{\newblock}{\relax}
\providecommand{\bibinfo}[2]{#2}
\providecommand{\BIBentrySTDinterwordspacing}{\spaceskip=0pt\relax}
\providecommand{\BIBentryALTinterwordstretchfactor}{4}
\providecommand{\BIBentryALTinterwordspacing}{\spaceskip=\fontdimen2\font plus
\BIBentryALTinterwordstretchfactor\fontdimen3\font minus \fontdimen4\font\relax}
\providecommand{\BIBforeignlanguage}[2]{{%
\expandafter\ifx\csname l@#1\endcsname\relax
\typeout{** WARNING: IEEEtran.bst: No hyphenation pattern has been}%
\typeout{** loaded for the language `#1'. Using the pattern for}%
\typeout{** the default language instead.}%
\else
\language=\csname l@#1\endcsname
\fi
#2}}
\providecommand{\BIBdecl}{\relax}
\BIBdecl

\bibitem{cheng2021causal}
L.~Cheng, R.~Guo, K.~Shu, and H.~Liu, ``Causal understanding of fake news dissemination on social media,'' in \emph{Proceedings of KDD}, 2021.

\bibitem{zafarani2019fake}
R.~Zafarani, X.~Zhou, K.~Shu, and H.~Liu, ``Fake news research: Theories, detection strategies, and open problems,'' in \emph{Proceedings of KDD}, 2019.

\bibitem{DBLP:journals/tois/YangSZLC17}
C.~Yang, M.~Sun, W.~X. Zhao, Z.~Liu, and E.~Y. Chang, ``A neural network approach to jointly modeling social networks and mobile trajectories,'' \emph{{ACM} Trans. Inf. Syst.}, 2017.

\bibitem{vosoughi2018spread}
S.~Vosoughi, D.~Roy, and S.~Aral, ``The spread of true and false news online,'' \emph{science}, no. 6380, 2018.

\bibitem{hassan2015detecting}
N.~Hassan, C.~Li, and M.~Tremayne, ``Detecting check-worthy factual claims in presidential debates,'' in \emph{Proceedings of CIKM}, 2015.

\bibitem{vlachos2014fact}
A.~Vlachos and S.~Riedel, ``Fact checking: Task definition and dataset construction,'' in \emph{Proceedings of the {ACL} 2014 Workshop on Language Technologies and Computational Social Science}, 2014.

\bibitem{guo2022survey}
Z.~Guo, M.~Schlichtkrull, and A.~Vlachos, ``A survey on automated fact-checking,'' \emph{Transactions of the Association for Computational Linguistics}, 2022.

\bibitem{wadden-etal-2020-fact}
D.~Wadden, S.~Lin, K.~Lo, L.~L. Wang, M.~van Zuylen, A.~Cohan, and H.~Hajishirzi, ``Fact or fiction: Verifying scientific claims,'' in \emph{Proceedings of EMNLP}, 2020.

\bibitem{DBLP:conf/emnlp/JiangBZD0B20}
Y.~Jiang, S.~Bordia, Z.~Zhong, C.~Dognin, M.~Singh, and M.~Bansal, ``{H}o{V}er: A dataset for many-hop fact extraction and claim verification,'' in \emph{Proceedings of EMNLP Findings}, 2020.

\bibitem{DBLP:conf/acl/ParkMKZH22}
J.~Park, S.~Min, J.~Kang, L.~Zettlemoyer, and H.~Hajishirzi, ``Faviq: Fact verification from information-seeking questions,'' in \emph{Proceedings of ACL}, 2022.

\bibitem{ma2019sentence}
J.~Ma, W.~Gao, S.~Joty, and K.-F. Wong, ``Sentence-level evidence embedding for claim verification with hierarchical attention networks,'' in \emph{Proceedings of ACL}, 2019.

\bibitem{wan2021dqn}
H.~Wan, H.~Chen, J.~Du, W.~Luo, and R.~Ye, ``A {DQN}-based approach to finding precise evidences for fact verification,'' in \emph{Proceedings of ACL}, 2021.

\bibitem{DBLP:journals/csur/BekoulisPD23}
G.~Bekoulis, C.~Papagiannopoulou, and N.~Deligiannis, ``A review on fact extraction and verification,'' \emph{{ACM} Comput. Surv.}, no.~2, 2023.

\bibitem{augenstein-etal-2019-multifc}
I.~Augenstein, C.~Lioma, D.~Wang, L.~Chaves~Lima, C.~Hansen, C.~Hansen, and J.~G. Simonsen, ``{M}ulti{FC}: A real-world multi-domain dataset for evidence-based fact checking of claims,'' in \emph{Proceedings of EMNLP}, 2019.

\bibitem{schuster2021get}
T.~Schuster, A.~Fisch, and R.~Barzilay, ``Get your vitamin {C}! robust fact verification with contrastive evidence,'' in \emph{Proceedings of NAACL-HLT}, 2021.

\bibitem{thorne2018fact}
J.~Thorne, A.~Vlachos, O.~Cocarascu, C.~Christodoulopoulos, and A.~Mittal, ``The fact extraction and {VER}ification ({FEVER}) shared task,'' in \emph{Proceedings of the First Workshop on Fact Extraction and {VER}ification ({FEVER})}, 2018.

\bibitem{schuster2019towards}
T.~Schuster, D.~Shah, Y.~J.~S. Yeo, D.~Roberto Filizzola~Ortiz, E.~Santus, and R.~Barzilay, ``Towards debiasing fact verification models,'' in \emph{Proceedings of EMNLP}, 2019.

\bibitem{kim-etal-2023-factkg}
J.~Kim, S.~Park, Y.~Kwon, Y.~Jo, J.~Thorne, and E.~Choi, ``{F}act{KG}: Fact verification via reasoning on knowledge graphs,'' in \emph{Proceedings of ACL}, 2023.

\bibitem{fajcik-etal-2023-claim}
M.~Fajcik, P.~Motlicek, and P.~Smrz, ``Claim-dissector: An interpretable fact-checking system with joint re-ranking and veracity prediction,'' in \emph{Proceedings of ACL Findings}, 2023.

\bibitem{yin2018twowingos}
W.~Yin and D.~Roth, ``{T}wo{W}ing{OS}: A two-wing optimization strategy for evidential claim verification,'' in \emph{Proceedings of EMNLP}, 2018.

\bibitem{chen2017reading}
D.~Chen, A.~Fisch, J.~Weston, and A.~Bordes, ``Reading {W}ikipedia to answer open-domain questions,'' in \emph{Proceedings of ACL}, 2017.

\bibitem{DBLP:conf/emnlp/Aly022}
R.~Aly and A.~Vlachos, ``Natural logic-guided autoregressive multi-hop document retrieval for fact verification,'' in \emph{Proceedings of EMNLP}, 2022.

\bibitem{nie2019combining}
Y.~Nie, H.~Chen, and M.~Bansal, ``Combining fact extraction and verification with neural semantic matching networks,'' in \emph{Proceedings of AAAI}, 2019.

\bibitem{hanselowski2018ukp}
A.~Hanselowski, H.~Zhang, Z.~Li, D.~Sorokin, B.~Schiller, C.~Schulz, and I.~Gurevych, ``{UKP}-athene: Multi-sentence textual entailment for claim verification,'' in \emph{Proceedings of the First Workshop on Fact Extraction and {VER}ification ({FEVER})}, 2018.

\bibitem{devlin2019bert}
J.~Devlin, M.-W. Chang, K.~Lee, and K.~Toutanova, ``{BERT}: Pre-training of deep bidirectional transformers for language understanding,'' in \emph{Proceedings of NAACL-HLT}, 2019.

\bibitem{DBLP:conf/acl/LiuXSL20}
Z.~Liu, C.~Xiong, M.~Sun, and Z.~Liu, ``Fine-grained fact verification with kernel graph attention network,'' in \emph{Proceedings of ACL}, 2020.

\bibitem{zhou2019gear}
J.~Zhou, X.~Han, C.~Yang, Z.~Liu, L.~Wang, C.~Li, and M.~Sun, ``{GEAR}: Graph-based evidence aggregating and reasoning for fact verification,'' in \emph{Proceedings of ACL}, 2019.

\bibitem{chen2022loren}
J.~Chen, Q.~Bao, C.~Sun, X.~Zhang, J.~Chen, H.~Zhou, Y.~Xiao, and L.~Li, ``Loren: Logic-regularized reasoning for interpretable fact verification,'' in \emph{Proceedings of AAAI}, no.~10, 2022.

\bibitem{DBLP:conf/coling/ParkLJKKN22}
E.~Park, J.~Lee, D.~H. Jeon, S.~Kim, I.~Kang, and S.~Na, ``{SISER:} semantic-infused selective graph reasoning for fact verification,'' in \emph{Proceedings of COLING}, 2022.

\bibitem{yoneda2018ucl}
T.~Yoneda, J.~Mitchell, J.~Welbl, P.~Stenetorp, and S.~Riedel, ``{UCL} machine reading group: Four factor framework for fact finding ({H}exa{F}),'' in \emph{Proceedings of the First Workshop on Fact Extraction and {VER}ification ({FEVER})}, 2018.

\bibitem{si2021topic}
J.~Si, D.~Zhou, T.~Li, X.~Shi, and Y.~He, ``Topic-aware evidence reasoning and stance-aware aggregation for fact verification,'' in \emph{Proceedings of ACL}, 2021.

\bibitem{velivckovic2018graph}
P.~Velickovic, G.~Cucurull, A.~Casanova, A.~Romero, P.~Li{\`{o}}, and Y.~Bengio, ``Graph attention networks,'' in \emph{Proceedings of ICLR}, 2018.

\bibitem{DBLP:conf/emnlp/WangZLZYZ22}
L.~Wang, P.~Zhang, X.~Lu, L.~Zhang, C.~Yan, and C.~Zhang, ``Qadialmoe: Question-answering dialogue based fact verification with mixture of experts,'' in \emph{Proceedings of EMNLP Findings}, 2022.

\bibitem{zhang-etal-2020-optimizing}
Y.~Zhang, D.~Merck, E.~Tsai, C.~D. Manning, and C.~Langlotz, ``Optimizing the factual correctness of a summary: A study of summarizing radiology reports,'' in \emph{Proceedings of ACL}, 2020.

\bibitem{parikh2016decomposable}
A.~Parikh, O.~T{\"a}ckstr{\"o}m, D.~Das, and J.~Uszkoreit, ``A decomposable attention model for natural language inference,'' in \emph{Proceedings of EMNLP}, 2016.

\bibitem{radford2018improving}
A.~Radford, K.~Narasimhan, T.~Salimans, and I.~Sutskever, ``Improving language understanding by generative pre-training,'' in \emph{Proceedings of Technical report, OpenAI}, 2018.

\bibitem{chen2017enhanced}
Q.~Chen, X.~Zhu, Z.-H. Ling, S.~Wei, H.~Jiang, and D.~Inkpen, ``Enhanced {LSTM} for natural language inference,'' in \emph{Proceedings of ACL}, 2017.

\bibitem{ghaeini2018dr}
R.~Ghaeini, S.~A. Hasan, V.~Datla, J.~Liu, K.~Lee, A.~Qadir, Y.~Ling, A.~Prakash, X.~Fern, and O.~Farri, ``{DR}-{B}i{LSTM}: Dependent reading bidirectional {LSTM} for natural language inference,'' in \emph{Proceedings of NAACL-HLT}, 2018.

\bibitem{peters2018deep}
M.~E. Peters, M.~Neumann, M.~Iyyer, M.~Gardner, C.~Clark, K.~Lee, and L.~Zettlemoyer, ``Deep contextualized word representations,'' in \emph{Proceedings of NAACL-HLT}, 2018.

\bibitem{li2019several}
T.~Li, X.~Zhu, Q.~Liu, Q.~Chen, Z.~Chen, and S.~Wei, ``Several experiments on investigating pretraining and knowledge-enhanced models for natural language inference,'' \emph{ArXiv preprint}, 2019.

\bibitem{hidey-diab-2018-team}
C.~Hidey and M.~Diab, ``Team {SWEEP}er: Joint sentence extraction and fact checking with pointer networks,'' in \emph{Proceedings of the First Workshop on Fact Extraction and {VER}ification ({FEVER})}, 2018.

\bibitem{chen-etal-2017-enhanced}
Q.~Chen, X.~Zhu, Z.-H. Ling, S.~Wei, H.~Jiang, and D.~Inkpen, ``Enhanced {LSTM} for natural language inference,'' in \emph{Proceedings of ACL}, 2017.

\bibitem{liu2019roberta}
Y.~Liu, M.~Ott, N.~Goyal, J.~Du, M.~Joshi, D.~Chen, O.~Levy, M.~Lewis, L.~Zettlemoyer, and V.~Stoyanov, ``Roberta: A robustly optimized bert pretraining approach,'' \emph{ArXiv preprint}, 2019.

\bibitem{lewis2020bart}
M.~Lewis, Y.~Liu, N.~Goyal, M.~Ghazvininejad, A.~Mohamed, O.~Levy, V.~Stoyanov, and L.~Zettlemoyer, ``{BART}: Denoising sequence-to-sequence pre-training for natural language generation, translation, and comprehension,'' in \emph{Proceedings of ACL}, 2020.

\bibitem{raffel2020exploring}
C.~Raffel, N.~Shazeer, A.~Roberts, K.~Lee, S.~Narang, M.~Matena, Y.~Zhou, W.~Li, and P.~J. Liu, ``Exploring the limits of transfer learning with a unified text-to-text transformer,'' \emph{Journal of Machine Learning Research}, 2020.

\bibitem{NEURIPS2019_dc6a7e65}
Z.~Yang, Z.~Dai, Y.~Yang, J.~G. Carbonell, R.~Salakhutdinov, and Q.~V. Le, ``Xlnet: Generalized autoregressive pretraining for language understanding,'' in \emph{Proceedings of NeurIPS}, 2019.

\bibitem{DBLP:conf/ecir/SoleimaniMW20}
A.~Soleimani, C.~Monz, and M.~Worring, ``{BERT} for evidence retrieval and claim verification,'' in \emph{Proc. of ECIR}, 2020.

\bibitem{jiang2021exploring}
K.~Jiang, R.~Pradeep, and J.~Lin, ``Exploring listwise evidence reasoning with t5 for fact verification,'' in \emph{Proceedings of ACL}, 2021.

\bibitem{lee2021towards}
N.~Lee, Y.~Bang, A.~Madotto, and P.~Fung, ``Towards few-shot fact-checking via perplexity,'' in \emph{Proceedings of NAACL-HLT}, 2021.

\bibitem{DBLP:conf/iclr/0002IWXJ000023}
W.~Yu, D.~Iter, S.~Wang, Y.~Xu, M.~Ju, S.~Sanyal, C.~Zhu, M.~Zeng, and M.~Jiang, ``Generate rather than retrieve: Large language models are strong context generators,'' in \emph{Proceedings of ICLR}, 2023.

\bibitem{DBLP:conf/nips/LewisPPPKGKLYR020}
P.~S.~H. Lewis, E.~Perez, A.~Piktus, F.~Petroni, V.~Karpukhin, N.~Goyal, H.~K{\"{u}}ttler, M.~Lewis, W.~Yih, T.~Rockt{\"{a}}schel, S.~Riedel, and D.~Kiela, ``Retrieval-augmented generation for knowledge-intensive {NLP} tasks,'' in \emph{Proceedings of NeurIPS}, 2020.

\bibitem{DBLP:conf/cikm/LeeWKLPJ21}
M.~Lee, S.~Won, J.~Kim, H.~Lee, C.~Park, and K.~Jung, ``Crossaug: {A} contrastive data augmentation method for debiasing fact verification models,'' in \emph{Proceedings of CIKM}, 2021.

\bibitem{pan-etal-2021-Zero-shot-FV}
L.~Pan, W.~Chen, W.~Xiong, M.-Y. Kan, and W.~Y. Wang, ``Zero-shot fact verification by claim generation,'' in \emph{Proceedings of ACL}, 2021.

\bibitem{10.1007/978-3-030-45442-5_45}
A.~Soleimani, C.~Monz, and M.~Worring, ``Bert for evidence retrieval and claim verification,'' in \emph{Advances in Information Retrieval}, 2020.

\bibitem{DBLP:conf/acl/KruengkraiYW21}
C.~Kruengkrai, J.~Yamagishi, and X.~Wang, ``A multi-level attention model for evidence-based fact checking,'' in \emph{Proceedings of ACL Findings}, 2021.

\bibitem{pradeep2021scientific}
R.~Pradeep, X.~Ma, R.~Nogueira, and J.~Lin, ``Scientific claim verification with {V}er{T}5erini,'' in \emph{Proceedings of the 12th International Workshop on Health Text Mining and Information Analysis}, 2021.

\bibitem{xu2023counterfactual}
W.~Xu, Q.~Liu, S.~Wu, and L.~Wang, ``Counterfactual debiasing for fact verification,'' in \emph{Proceedings of ACL}, 2023.

\bibitem{DBLP:journals/tacl/Krishna0022}
A.~Krishna, S.~Riedel, and A.~Vlachos, ``Proofver: Natural logic theorem proving for fact verification,'' \emph{Trans. Assoc. Comput. Linguistics}, 2022.

\bibitem{Chen-evidencenet}
Z.~Chen, S.~C. Hui, F.~Zhuang, L.~Liao, F.~Li, M.~Jia, and J.~Li, ``Evidencenet: Evidence fusion network for fact verification,'' in \emph{Proceedings of the ACM Web Conference 2022}, 2022.

\bibitem{wang-etal-2022-imci}
H.~Wang, Y.~Li, Z.~Huang, and Y.~Dou, ``{IMCI}: Integrate multi-view contextual information for fact extraction and verification,'' in \emph{Proceedings of COLING}, 2022.

\bibitem{barik2022incorporating}
A.~M. Barik, W.~Hsu, and M.~L. Lee, ``Incorporating external knowledge for evidence-based fact verification,'' in \emph{Companion Proceedings of the Web Conference 2022}, 2022.

\bibitem{zhong-etal-2020-reasoning}
W.~Zhong, J.~Xu, D.~Tang, Z.~Xu, N.~Duan, M.~Zhou, J.~Wang, and J.~Yin, ``Reasoning over semantic-level graph for fact checking,'' in \emph{Proceedings of ACL}, 2020.

\bibitem{kipf2016semi}
T.~N. Kipf and M.~Welling, ``Semi-supervised classification with graph convolutional networks,'' in \emph{Proceedings of ICLR}, 2017.

\bibitem{DBLP:conf/coling/MaL0C22}
Z.~Ma, J.~Li, G.~Li, and Y.~Cheng, ``{GLAF:} global-to-local aggregation and fission network for semantic level fact verification,'' in \emph{Proceedings of COLING}, 2022.

\bibitem{chen2021entity}
C.~Chen, F.~Cai, X.~Hu, J.~Zheng, Y.~Ling, and H.~Chen, ``An entity-graph based reasoning method for fact verification,'' \emph{Information Processing \& Management}, no.~3, 2021.

\bibitem{clark2020electra}
K.~Clark, M.~Luong, Q.~V. Le, and C.~D. Manning, ``{ELECTRA:} pre-training text encoders as discriminators rather than generators,'' in \emph{Proceedings of ICLR}, 2020.

\bibitem{vaswani2017attention}
A.~Vaswani, N.~Shazeer, N.~Parmar, J.~Uszkoreit, L.~Jones, A.~N. Gomez, L.~Kaiser, and I.~Polosukhin, ``Attention is all you need,'' in \emph{Proceedings of NeurIPS}, 2017.

\bibitem{clark-etal-2019-bert}
K.~Clark, U.~Khandelwal, O.~Levy, and C.~D. Manning, ``What does {BERT} look at? an analysis of {BERT}{'}s attention,'' in \emph{Proceedings of the 2019 ACL Workshop BlackboxNLP: Analyzing and Interpreting Neural Networks for NLP}, 2019.

\bibitem{brody2021attentive}
S.~Brody, U.~Alon, and E.~Yahav, ``How attentive are graph attention networks?'' in \emph{Proceedings of ICLR}, 2021.

\bibitem{thorne2018fever}
J.~Thorne, A.~Vlachos, C.~Christodoulopoulos, and A.~Mittal, ``{FEVER}: a large-scale dataset for fact extraction and {VER}ification,'' in \emph{Proceedings of NAACL-HLT}, 2018.

\bibitem{liu2020adapting}
Z.~Liu, C.~Xiong, Z.~Dai, S.~Sun, M.~Sun, and Z.~Liu, ``Adapting open domain fact extraction and verification to {COVID}-{FACT} through in-domain language modeling,'' in \emph{Proceedings of EMNLP Findings}, 2020.

\bibitem{ye2020coreferential}
D.~Ye, Y.~Lin, J.~Du, Z.~Liu, P.~Li, M.~Sun, and Z.~Liu, ``{C}oreferential {R}easoning {L}earning for {L}anguage {R}epresentation,'' in \emph{Proceedings of EMNLP}, 2020.

\bibitem{gardner2018allennlp}
M.~Gardner, J.~Grus, M.~Neumann, O.~Tafjord, P.~Dasigi, N.~F. Liu, M.~Peters, M.~Schmitz, and L.~Zettlemoyer, ``{A}llen{NLP}: A deep semantic natural language processing platform,'' in \emph{Proceedings of Workshop for {NLP} Open Source Software ({NLP}-{OSS})}, 2018.

\bibitem{DBLP:conf/emnlp/WangLWLS19}
S.~Wang, Y.~Liu, C.~Wang, H.~Luan, and M.~Sun, ``Improving back-translation with uncertainty-based confidence estimation,'' in \emph{Proceedings of EMNLP}, 2019.

\bibitem{DBLP:conf/icml/OttAGR18}
M.~Ott, M.~Auli, D.~Grangier, and M.~Ranzato, ``Analyzing uncertainty in neural machine translation,'' in \emph{Proceedings of ICML}, 2018.

\bibitem{ni-etal-2022-sentence}
J.~Ni, G.~Hernandez~Abrego, N.~Constant, J.~Ma, K.~Hall, D.~Cer, and Y.~Yang, ``Sentence-t5: Scalable sentence encoders from pre-trained text-to-text models,'' in \emph{Proceedings of ACL Findings}, 2022.

\end{thebibliography}
\begin{IEEEbiography}[{\includegraphics[width=1in,height=1.25in,clip,keepaspectratio]{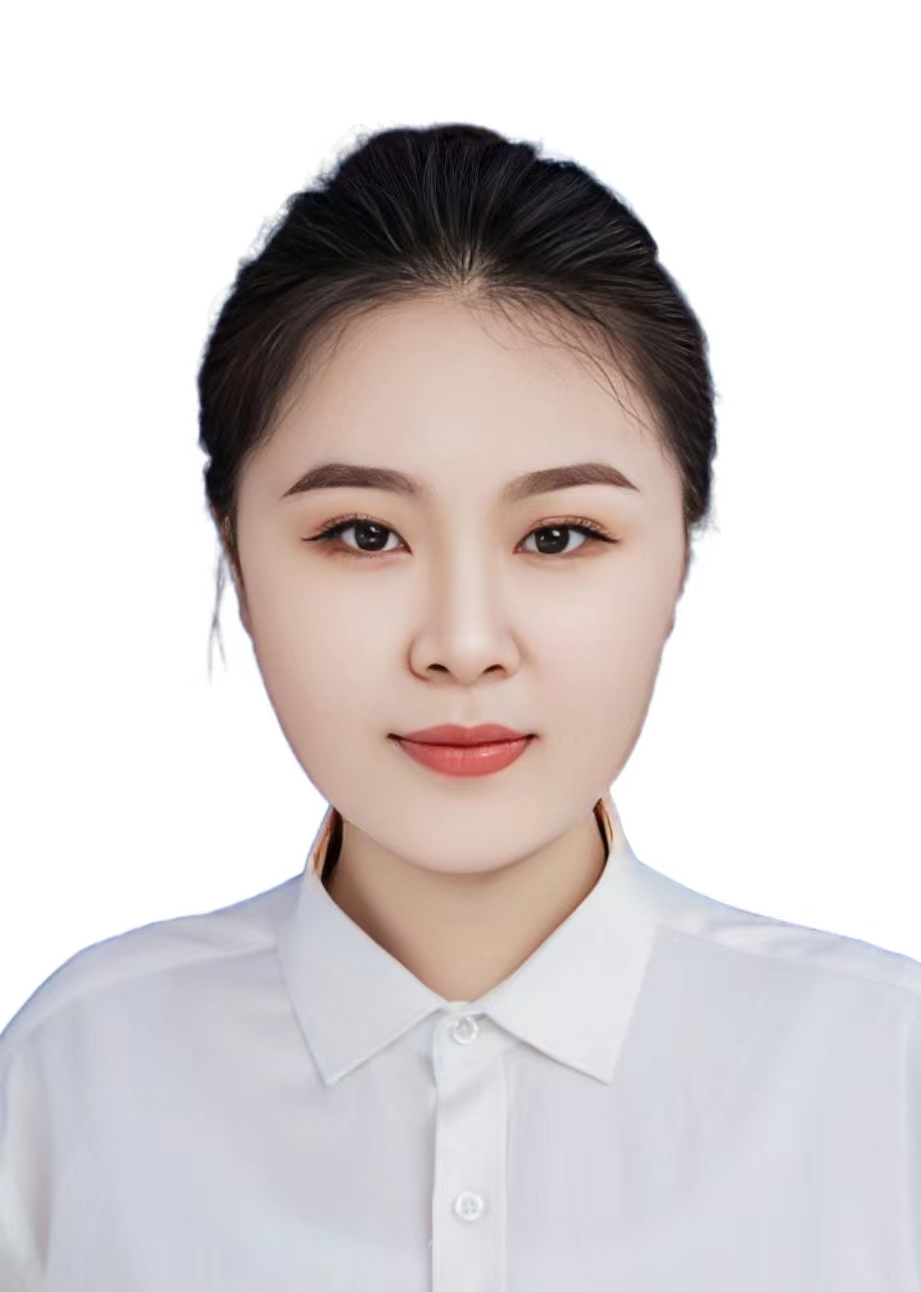}}]{Yuqing Lan}
received the M.S. degree in computer Application Technology from Northeastern University, China, in 2020. She is currently working toward her Ph.D. degree in the School of Computer Science and Engineering of Northeastern University. Her current research interests include fact verification.\end{IEEEbiography}
\begin{IEEEbiography}[{\includegraphics[width=1in,height=1.25in,clip,keepaspectratio]{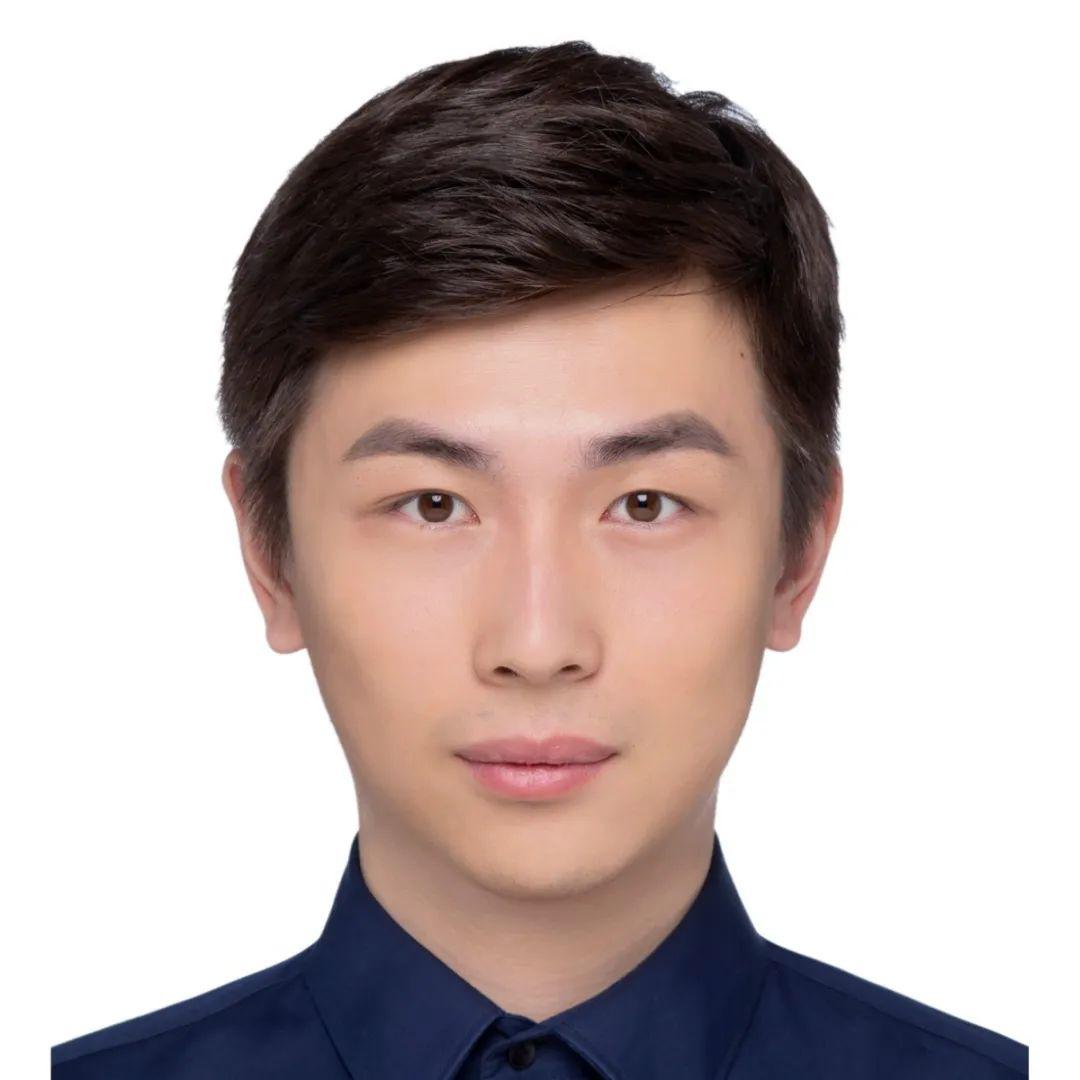}}]{Zhenghao Liu} received the Ph.D. degree in computer science and technology from Tsinghua University, China, in 2021. He is currently an associate professor at Northeastern University, China. His current research interests include natural language processing and information retrieval.
\end{IEEEbiography}

\begin{IEEEbiography}[{\includegraphics[width=1in,height=1.25in,clip,keepaspectratio]{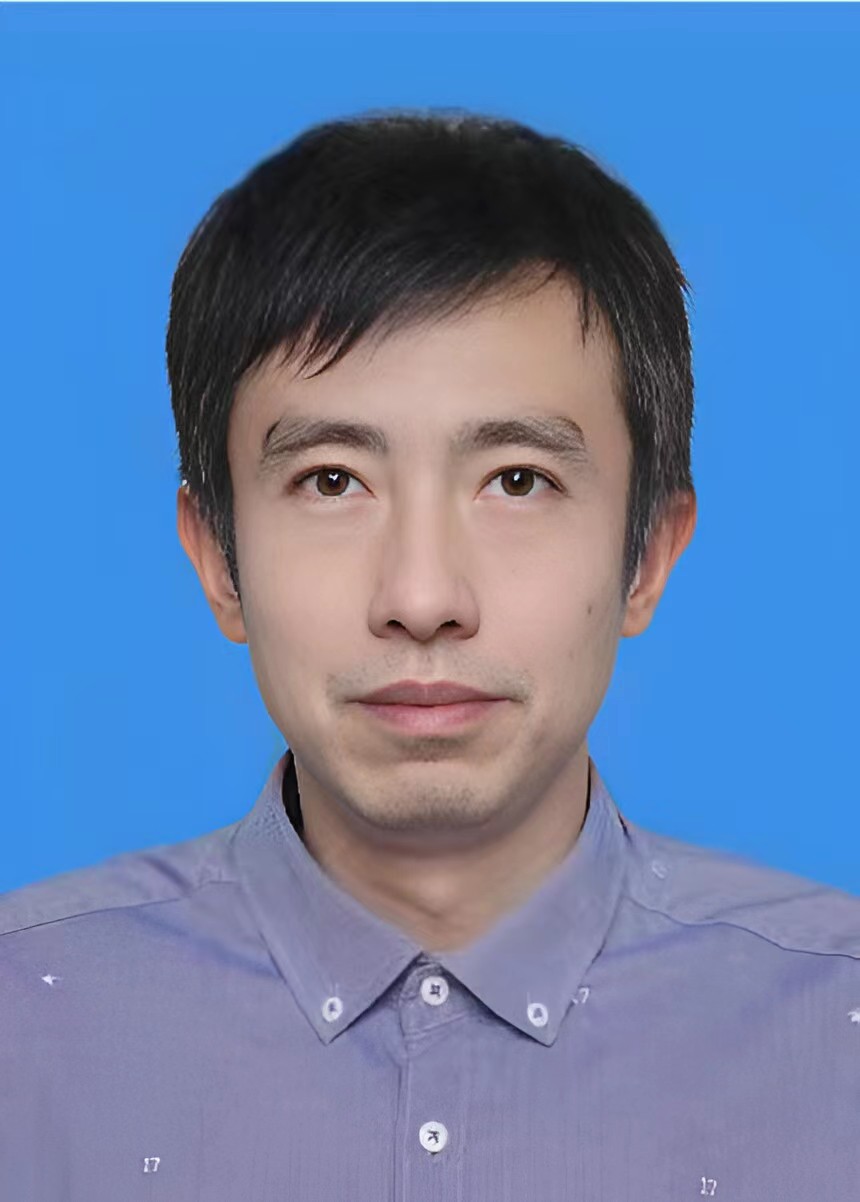}}]{Yu Gu}
received the Ph.D. degree in computer
software and theory from Northeastern University, China, in 2010. He is currently a professor
at Northeastern University, China. His current
research interests include big data processing and graph data management. He is a senior member of China Computer Federation (CCF).\end{IEEEbiography}
\begin{IEEEbiography}[{\includegraphics[width=1in,height=1.25in,clip,keepaspectratio]{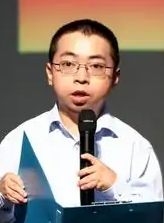}}]{Xiaoyuan Yi}
received the Ph.D. degree in
computer software and theory from Tsinghua
University, China, in 2021. He is currently at Microsoft Research Asia, Beijing, China. His current
research interests Natural Language Generation (NLG).\end{IEEEbiography}
\begin{IEEEbiography}[{\includegraphics[width=1in,height=1.25in,clip,keepaspectratio]{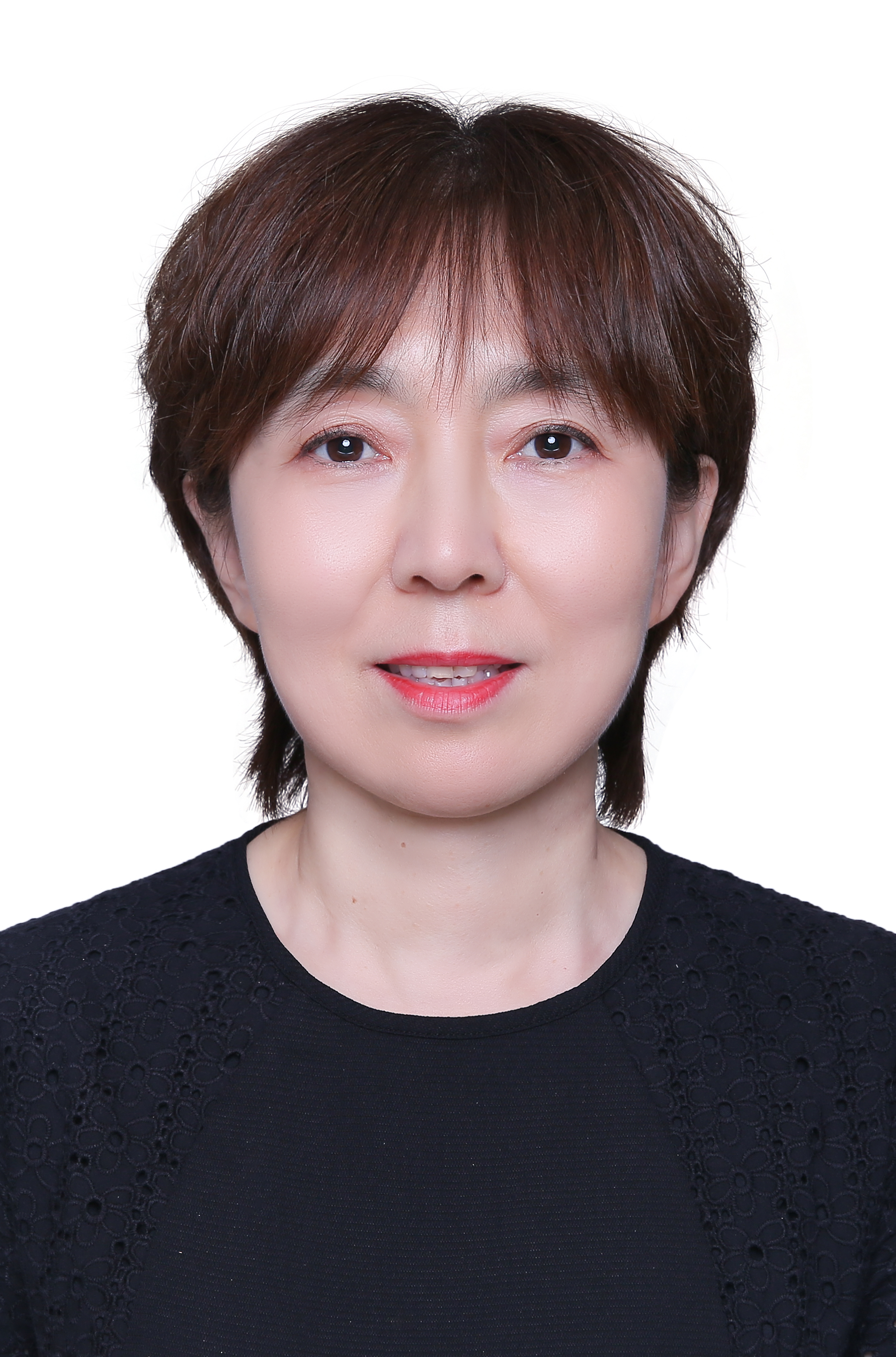}}]{Xiaohua Li}
received the Ph.D. degree in computer
software and theory from Northeastern University, China, in 2018. She is currently an associate professor
at Northeastern University, China. Her current
research interests include information security and block chain.\end{IEEEbiography}
\begin{IEEEbiography}[{\includegraphics[width=1in,height=1.25in,clip,keepaspectratio]{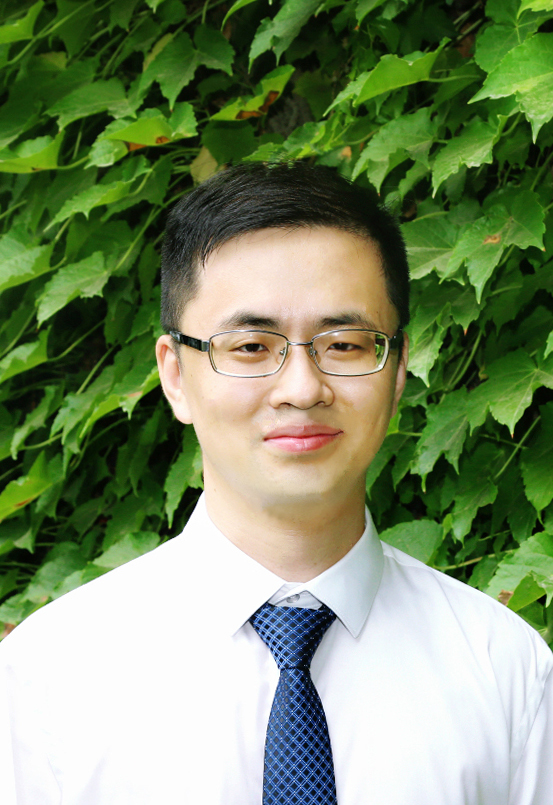}}]{Liner Yang}
received the Ph.D. degree in computer science from Tsinghua University, Beijing, China, in 2018. He is currently an associate professor at the National Language Resources Monitoring and Research Center for Print Media, Beijing Language and Culture University, Beijing, China. 
His current research interests include artificial intelligence, natural language processing, and NLP for educational applications.\end{IEEEbiography}

\begin{IEEEbiography}[{\includegraphics[width=0.9in,height=1.25in,clip,keepaspectratio]{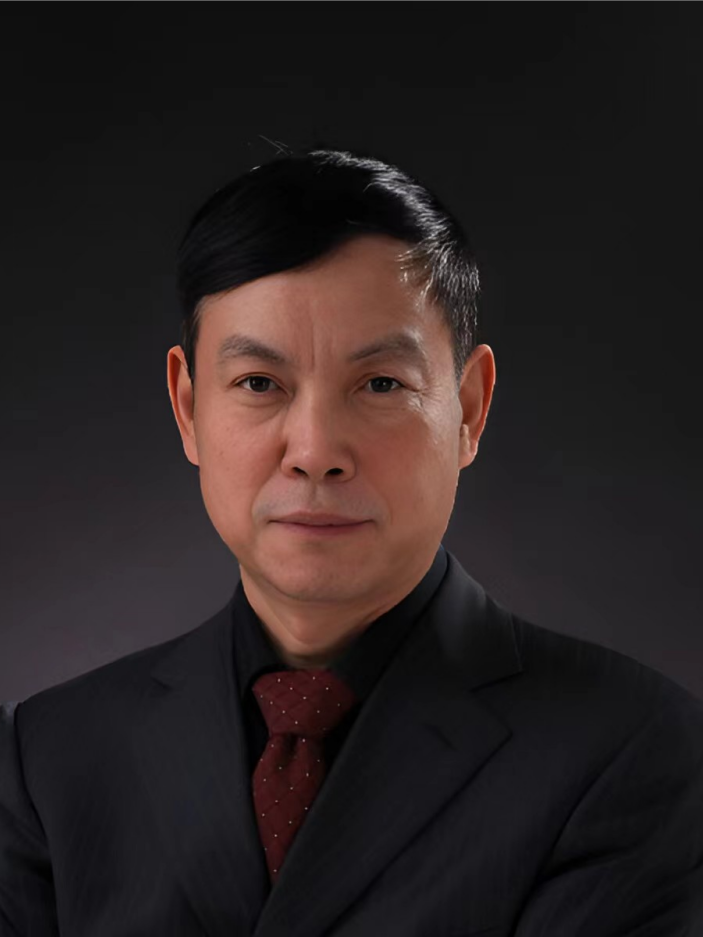}}]{Ge Yu}
received the Ph.D. degree in computer
science from the Kyushu University of Japan, in
1996. He is currently a professor at the Northeastern University of China. His research interests include distributed and parallel database,
data integration, and graph data management.
He is a fellow of CCF and a member of the IEEE
and ACM\end{IEEEbiography}

\vfill
\end{document}